\documentclass[10pt,twocolumn,letterpaper]{article}

\usepackage{cvpr}              %
\usepackage{algorithm}
\usepackage{algpseudocode}

\usepackage{multirow}
\usepackage{amsmath}
\usepackage{amssymb}
\usepackage{booktabs}

\usepackage[dvipsnames]{xcolor}

\newcounter{alphasect}
\def\alphainsection{0}

\let\oldsection=\section
\def\section{%
  \ifnum\alphainsection=1%
    \addtocounter{alphasect}{1}
  \fi%
\oldsection}%

\renewcommand\thesection{%
  \ifnum\alphainsection=1%
    \Alph{alphasect}
  \else%
    \arabic{section}
  \fi%
}%

\newenvironment{alphasection}{%
  \ifnum\alphainsection=1%
    \errhelp={Let other blocks end at the beginning of the next block.}
    \errmessage{Nested Alpha section not allowed}
  \fi%
  \setcounter{alphasect}{0}
  \def\alphainsection{1}
}{%
  \setcounter{alphasect}{0}
  \def\alphainsection{0}
}%

\definecolor{cvprblue}{rgb}{0.21,0.49,0.74}
\usepackage[pagebackref,breaklinks,colorlinks,citecolor=cvprblue]{hyperref}

\def\paperID{6159} %
\def\confName{CVPR}
\def\confYear{2024}

\begin{document}
\title{CyberDemo: Augmenting Simulated Human Demonstration \\
for Real-World Dexterous Manipulation}

\author{
Jun Wang\textsuperscript{1*} \quad 
Yuzhe Qin\textsuperscript{1*} \quad 
Kaiming Kuang\textsuperscript{1} \quad
Yigit Korkmaz\textsuperscript{2} \quad
Akhilan Gurumoorthy\textsuperscript{1} \quad \\
Hao Su\textsuperscript{1} \quad
Xiaolong Wang\textsuperscript{1} \quad \\
\textsuperscript{1}UC San Diego \quad
\textsuperscript{2}University of Southern California
}

\newcommand{\hao}[1]{{\color{red}[hao: #1]}}

\twocolumn[{%
\renewcommand\twocolumn[1][]{#1}%
\maketitle
\vspace{-3em}
\begin{center}
    \centering
    \captionsetup{type=figure}
    \includegraphics[width=\linewidth]{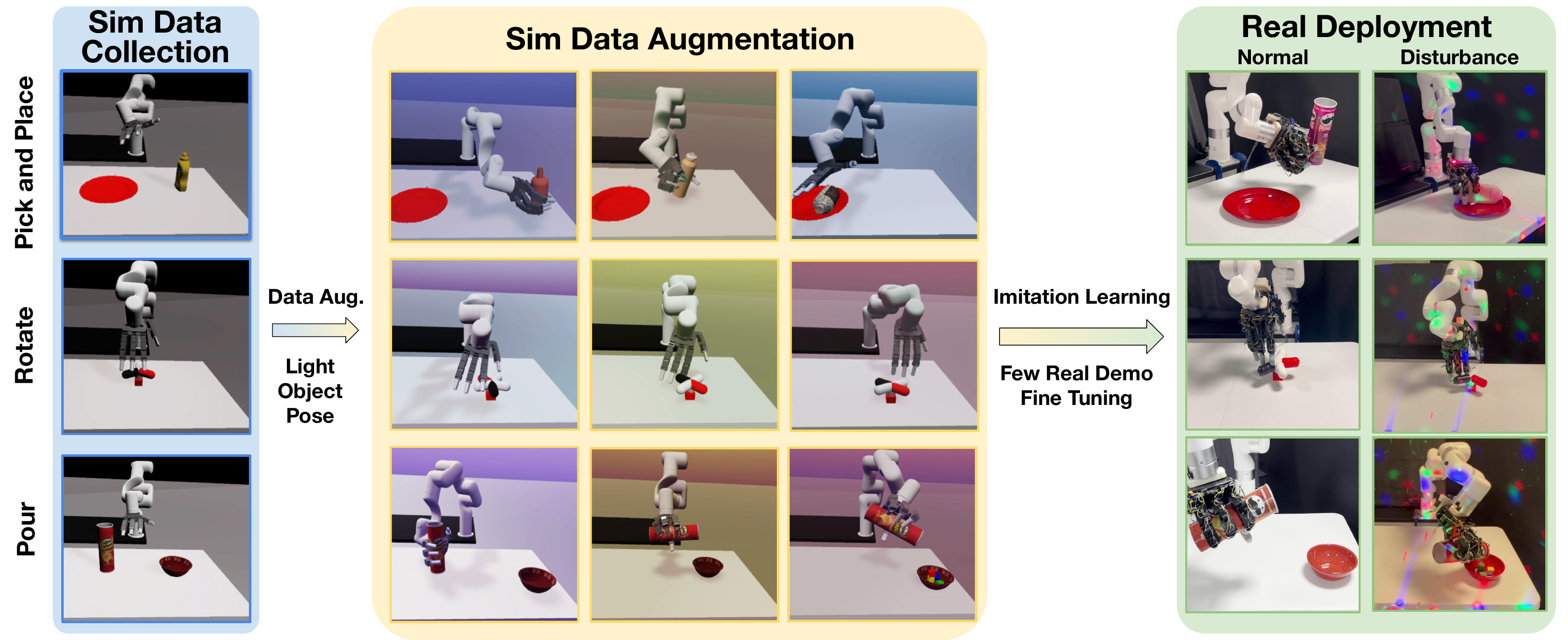}
    \vspace{-1em}
    \captionof{figure}{We propose CyberDemo, a novel pipeline for learning real-world dexterous manipulation by using simulation data. First, we collect human demos in a simulated environment (blue region), followed by extensive data augmentation within the simulator (yellow region). Then, the imitation learning model, trained on augmented data and fine-tuned on a few real data, can be deployed on a real robot.}
    \label{fig:teaser}
\end{center}
}]
\def\thefootnote{*}\footnotetext{Equal contributions.}

\begin{abstract}

\vspace{-0.8em}
We introduce \textbf{CyberDemo}, a novel approach to robotic imitation learning that leverages simulated human demonstrations for real-world tasks. By incorporating extensive data augmentation in a simulated environment, CyberDemo outperforms traditional in-domain real-world demonstrations when transferred to the real world, handling diverse physical and visual conditions. Regardless of its affordability and convenience in data collection, CyberDemo outperforms baseline methods in terms of success rates across various tasks and exhibits generalizability with previously unseen objects. For example, it can rotate novel tetra-valve and penta-valve, despite human demonstrations only involving tri-valves. Our research demonstrates the significant potential of simulated human demonstrations for real-world dexterous manipulation tasks. More
details can be found at \href{https://cyber-demo.github.io/}{https://cyber-demo.github.io/}

\end{abstract}
    
\section{Introduction}
\label{sec:intro}
Imitation learning has been a promising approach in robot manipulation, facilitating the acquisition of complex skills from human demonstration. However, the effectiveness of this approach is critically dependent on the availability of high-quality demonstration data, which often necessitates substantial human effort for data collection~\cite{rt-1, rt-x, roboagent}. This challenge is further amplified in the context of manipulation with a multi-finger dexterous hand, where the complexity and intricacy of the tasks require highly detailed and precise demonstrations.

In imitation learning, in-domain demonstrations, which refer to the data collected directly from the deployment environment, are commonly used for robot manipulation tasks~\cite{robo-mimic}. It is generally believed that the most effective way to solve a specific task is to collect demonstrations directly from the real robot on that task. This belief has been upheld as the gold standard, but we wish to challenge it. We argue that collecting human demonstrations in simulation can yield superior results for real-world tasks, not only because it does not require real hardware and can be executed remotely and in parallel, but also due to its potential to enhance final task performance by employing simulator-only data augmentation~\cite{roboturk, kar2019meta, savva2019habitat, behavior1k, from-one-hand, mimicgen}.
This allows the generation of a dataset that is hundreds of times larger than the initial demonstration set. However, while existing studies employ the generated dataset to train in-domain policies within the simulation, the sim2real challenge of transferring policies to the real world remains an unresolved problem.

In this paper, we study the problem of how to utilize simulated human demos for real-world robot manipulation tasks.
We introduce \textbf{CyberDemo}, a novel framework designed for robotic imitation learning from visual observations, leveraging simulated human demos. We first collect a modest amount of human demonstration data via teleoperation using low-cost devices in a simulated environment. Then, CyberDemo incorporates extensive data augmentation into the original human demonstration. 
The augmented set covers a broad spectrum of visual and physical conditions not encountered during data collection, thereby enhancing the robustness of the trained policy against these variations. These augmentation techniques are also designed with the downstream sim2real transfer in mind.
We employ a unique curriculum learning strategy to train the policy on the augmented dataset, then fine-tune it using a few real-world demos (3-minute trajectories), facilitating effective transfer to real-world conditions.
While policies trained on only real-world demonstrations may suffer from variations in lighting conditions, object geometry, and object initial pose, our policy is capable of handling these without the need for additional human effort.

Our system, which utilizes a low-cost motion capture device for teleoperation (i.e., RealSense camera) and demands minimal human effort (i.e., a 30-minute demo trajectory), can learn a robust imitation learning policy. Despite its affordability and minimal human effort requirements, CyberDemo can still achieve better performance on the real robot. Compared with pre-trained policies, e.g. R3M~\cite{r3m} fine-tuned on real-world demonstrations, CyberDemo achieves a success rate that is $35\%$ higher for quasi-static \textit{pick and place} tasks, and $20\%$ higher for non-quasi-static \textit{rotate} tasks. 
In the generalization test, while baseline methods struggle to handle unseen objects during testing, our method can rotate novel tetra-valve and penta-valve with $42.5\%$ success rate, even though human demonstrations only cover tri-valve (second row of Figure~\ref{fig:teaser}). Our method can also manage significant light disturbances (last column of Figure~\ref{fig:teaser}).
In our ablation study, we observe that the use of data augmentation, coupled with an increased number of demonstrations in the simulator, results in superior performance compared to an equivalent increase in real-world demonstrations. To foster further research, we will make our code and human demonstration dataset publicly available.

\section{Related Work}
\label{sec:related}

\textbf{Data for Learning Robot Manipulation.}
Imitation learning has been proven to be an effective approach to robotic manipulation, enabling policy training with a collection of demonstrations. 
Many works have focused on building large datasets using pre-programmed policies~\cite{rlbench, transporter, dalal2023imitating, maniskill2, vima}, alternative data sources such as language~\cite{huang2023voxposer, stepputtis2020language, progprompt, cliport} and human video~\cite{dexmv, videodex, nguyen2018translating, bharadhwaj2023visual, concept2robot} or extensive real-world robot teleoperation~\cite{rt-1, rt-x,  arunachalam2023holo, arunachalam2023dexterous, roboagent, bc-z, bridge-data, interactive-language, robo-mimic}. However, such works predominantly targeted parallel grippers. Collecting large-scale demonstration datasets for high-DoF dexterous hands continues to be a significant challenge.
Meanwhile, data augmentation presents a viable strategy to improve policy generalization by increasing the diversity of data distribution. Previous studies have applied augmentation in low-level visual space~\cite{rl-cyclegan, retinagan, cubuk1805autoaugment, shorten2019survey}, such as color jitter, blurring, and cropping, while more recent works propose semantic-aware data augmentation with generative models~\cite{yu2023scaling, cacti, genaug, black2023zero, chen2022learning, zhu2023diffusion}. However, these augmentations operate at the image level and are not grounded in physical reality. CyberDemo extends data augmentation to the trajectory level using a physical simulator, accounting for both visual and physical variations.
Concurrent to our work, MimicGen~\cite{mimicgen} proposes a system to synthesize demonstrations for long-horizon tasks by integrating multiple human trajectories. However, it confines demonstrations to in-domain learning, i.e., it only trains simulation policies with simulated demos without transferring to real robots.
In contrast, our work aims to harness simulation for real-world problem-solving. We exploit the convenience of simulators for collecting robot demonstrations and employ a sim2real approach to transfer these demos to a dexterous robot equipped with a multi-finger humanoid hand. Our research emphasizes a general framework that leverages simulated demonstrations for real-world robot manipulation.

\begin{figure*}[t]
  \centering
  \includegraphics[width=0.98\linewidth]{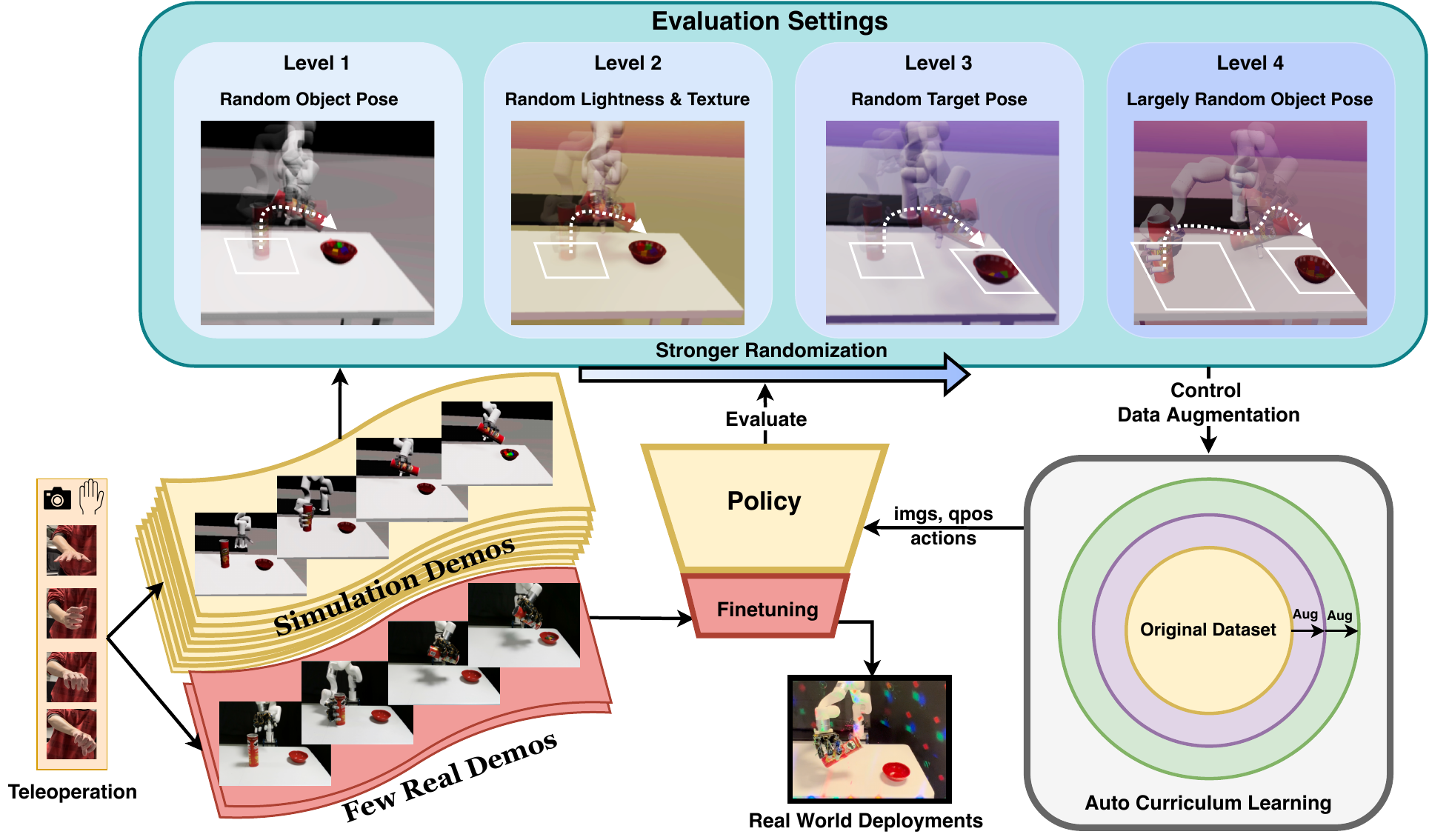}
   \caption{\textbf{CyberDemo Pipeline.} First, we collect both simulated and real demonstrations via vision-based teleoperation. Following this, we train the policy on simulated data, incorporating the proposed data augmentation techniques. During training, we apply automatic curriculum learning, which incrementally enhances the randomness scale based on task performance. Finally, the policy is fine-tuned with a few real demos before being deployed to the real world.}
   \label{fig:pipeline}
   \vspace{-1em}
\end{figure*}

\noindent
\textbf{Pre-trained Visual Representation for Robotics}
Recent progress in large-scale Self-Supervised Learning~\cite{moco, dino, mae} has enabled the development of visual representations that are advantageous for downstream robotic tasks~\cite{shah2021rrl, ze2023visual, yuan2022pre}. Several studies have focused on pretraining on non-robotic datasets, such as ImageNet~\cite{imagenet} and Ego4D~\cite{ego4d}, and utilizing the static representations for downstream robot control~\cite{r3m, pvr, mvp}.
Other research has focused on pre-training visual representations on robot datasets, using action-supervised self-learning objectives that depend on actions~\cite{videodex, schwarzer2021pretraining}, or utilizing the temporal consistency of video as a learning objective~\cite{yen2020learning, schmeckpeper2020reinforcement, stooke2021decoupling, tcn}. These investigations primarily aimed to learn features for effective training of vision-based robotic manipulation.
In addition to training visual representations on offline datasets, some researchers have also explored learning the reward function to be used in reinforcement learning~\cite{xirl, aytar2018playing, mees2020adversarial, dexvip, kumar2023graph}.
Unlike prior studies, our work diverges by utilizing simulation data for pre-training rather than employing Self-Supervised Learning for representation learning. This not only enhances the learning of image representations but also incorporates task priors into the neural network through the use of action information.
By pre-training in simulated environments, the manipulation policy can better generalize to new objects with novel geometries and contact patterns.

\noindent
\textbf{Sim2Real Transfer}
The challenge of transferring skills from simulation to real-world scenarios, known as sim2real transfer, has been a key focus in robot learning. Some approaches have employed system identification to build a mathematical model of real systems and identify physical parameters~\cite{kaspar2020sim2real, huang2023dynamic, chen2023visual, qi2023general, liang2020learning, real2sim2real}. Instead of calibrating real-world dynamics, domain randomization~\cite{domain-random, peng2018sim} generates simulated environments with randomized properties and trains a model function across all of them. Subsequent research demonstrated that the selection of randomization parameters could be automated~\cite{dextreme, akkaya2019solving, zakharov2019deceptionnet, chebotar2019closing}. However, due to the extensive sample requirements to learn robust policies, domain randomization is typically used with RL involving millions of interaction samples.
Domain adaptation (DA) refers to a set of transfer learning strategies developed to align the data distribution between sim and real. Common techniques include domain adversarial training~\cite{dann, truong2021bi} and the use of generative models to make simulated images resemble real ones~\cite{retinagan, bousmalis2017unsupervised}. 
Most of these DA approaches focus on bridging the visual gap. However, the challenge of addressing the dynamics gap remains significant. The sim2real gap becomes even more pronounced for dexterous robotic hands that have high-DoF actuation and complex interaction patterns~\cite{dextreme, rotating-touch, hora, yuan2023robot}.
In this work, we extend the concept of domain randomization to human demonstration collected in the simulator and focus on data augmentation techniques that can effectively utilize the simulation for transfer to a real robot. We demonstrate that there
can be a significant benefit in collecting human demonstration in the simulator, despite the sim2real gap, instead of solely relying on real data.

\section{CyberDemo}
\label{sec:method}

In CyberDemo, we initially gather human demonstrations of the same task in a simulator through teleoperation (Section~\ref{sec:teleop}). Taking advantage of the simulator's sampling capabilities and oracle state information, we enhance the simulated demonstration in various ways, increasing its visual, kinematic, and geometric diversity, thereby enriching the simulated dataset (Section~\ref{sec:aug}). With this augmented dataset, we train a manipulation policy with Automatic Curriculum Learning and Action Aggregation (Section~\ref{sec:learning}).

\subsection{Collecting Human Teleoperation Data}
\label{sec:teleop}

For each dexterous manipulation task in this work, we collect human demonstrations using teleoperation in both simulated and real-world environments.
For real-world data, we utilize the low-cost teleoperation system referenced in ~\cite{anyteleop}. This vision-based teleoperation system solely needs a camera to capture human hand motions as input, which are then translated into real-time motor commands for the robot arm and the dexterous hand. We record the observation (RGB image, robot proprioception) and the action (6D Cartesian velocity of robot end effector, finger joint position control target) for each frame at a rate of 30Hz. For this work, we collect only three minutes of robot trajectories for each task on the real robot.

For data in simulation, we build the real-world task environments within the SAPIEN~\cite{sapien} simulator to replicate the tables and objects used in real scenarios. It is worth noting that, for teleoperation, there is no requirement of reward design and observation spaces as in reinforcement learning settings, making the process of setting up new tasks in the simulator relatively simple. We employ the same teleoperation system~\cite{anyteleop} to collect human demonstrations in the simulator.

\begin{figure}[t]
  \centering
  \includegraphics[width=1\linewidth]{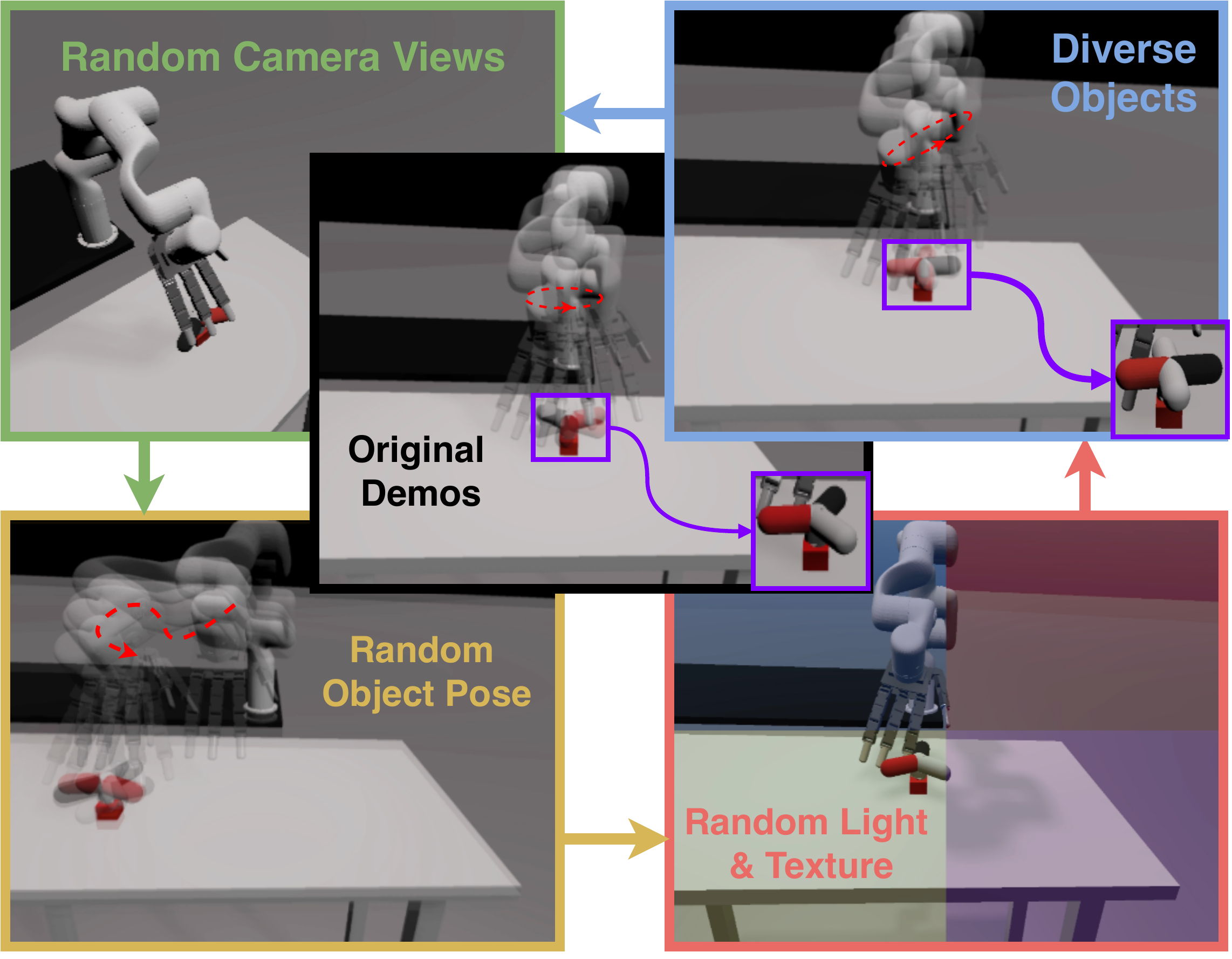}
   \caption{\textbf{Data Augmentation}. Our dataset augmentation encompasses four dimensions: (a) random camera views, (b) diverse objects, (c) random object pose, (d) random light and texture.}
   \label{fig:augmentation}
\end{figure}

\subsection{Augmenting Human Demo in Simulator}
\label{sec:aug}
Unlike real-world data collection, where we are limited to recording observations of physical sensors, such as camera RGB images and robot proprioception, the simulation system enables us to record the ground-truth state and contact information within the virtual environment. This unique benefit of simulation provides a more comprehensive data format for the simulated demonstrations compared to its real-world counterparts. Thus, we can take advantage of demonstration replay techniques on these simulated demonstrations, which are not feasible with real-world data.

When developing data augmentation techniques in the simulator, it is essential to keep in mind that the ultimate goal is to deploy the trained policy to a real robot. The augmentation should accordingly focus on the visual and dynamical variations that are likely to be encountered in the real world. 
Moreover, we aim for the manipulation policy to generalize to novel objects not encountered during the data collection process. For example, manipulating the tetra-valve when collecting data only on the tri-valve in Figure~\ref{fig:augmentation}.
Specifically, we chose to augment the lighting conditions, camera views, and object textures to enhance the policy's robustness against visual variations. In addition, we modified the geometric shape of the objects and the initial poses of the robot and objects to improve the policy's robustness against dynamical variations as follows:

\noindent \textbf{Randomize Camera Views.}
Precisely aligning camera views between demo collection and final evaluation, not to mention between simulation and reality, poses a significant challenge. To solve this problem, we randomize the camera pose during training and replay the internal state of the simulator to render image sequences from new camera views. Unlike standard image augmentation techniques such as cropping and shifting, our method respects the perspective projection in a physically realistic manner.

\noindent \textbf{Random Light and Texture.} 
To facilitate sim2real transfer and improve the policy's robustness against visual variations, we randomize the visual properties of both lights and objects (Figure~\ref{fig:augmentation}, lower right). Light properties include directions, colors, shadow characteristics, and ambient illumination. Object properties include specularity, roughness, metallicity, and texture. Similar to camera view randomization, we can simply replay the simulation state to render new image sequences.

\noindent \textbf{Add Diverse Objects.} 
In this approach, we replace the manipulated object in the original demos with novel objects (Figure~\ref{fig:augmentation} upper right). However, directly replaying the same trajectory would not work as the object shape is different. Instead, we perturb the action sequence from the original demo with Gaussian noises to generate new trajectories. These trajectories provide reasonable manipulation strategies but are slightly different from the original one. With the highly cost-effective sampling in the simulator, we can enumerate the perturbation until it is successful. It is important to note that this technique is feasible with real-world demonstrations.

\noindent \textbf{Randomize Object Pose.} 
A common approach in reinforcement learning to enhance generalizability involves randomizing the object pose during reset. Augmenting imitation learning data to achieve a similar outcome, however, is less intuitive.
Denote $T_{A}^{B}\in SE(3)$ as the pose of frame $B$ relative to frame $A$. The original object pose is $T_{W}^{O_{old}}$, the newly randomized object pose is $T_{W}^{O_{new}}$, and the original end effector pose is $T_{W}^{R_{old}}$. The objective is to handle the object pose change $T_{W}^{O_{new}} (T_{W}^{O_{old}})^{-1}$. A simple strategy can be first moving the robot end effector to a new initial pose, $T_{W}^{R_{new}} = T_{W}^{O_{new}} (T_{W}^{O_{old}})^{-1} T_{W}^{R_{old}}$. Then, the relative pose between the robot and the object aligns with the original demonstration, enabling us to replay the same action sequence to accomplish the task. Although this method succeeds in generating new trajectories, it offers minimal assistance for downstream imitation learning. The new trajectory is always composed of two segments: a computed reaching trajectory to the new end effector pose $T_{W}^{R_{new}}$, and the original trajectory. Given that different augmented trajectories often share a significant portion of redundancy, they fail to provide substantial new information to learning algorithms. 

\renewcommand{\algorithmicrequire}{\textbf{Input:}}
\renewcommand{\algorithmicensure}{\textbf{Output:}}
\newcommand{\squad}{\mkern9mu}

\begin{algorithm}[t]
\caption{Auto Curriculum Learning}\label{alg:adr}

\begin{algorithmic}[1]
\Require human demo $D_h$, training set $D$, policy network $\pi$, curriculum level $L$,  evaluation function $\mathbf{eval}_{L}()$, \quad data aug. function $\mathbf{aug}_{L}()$, success rate threshold $r_{up}$, number of failure $N_{fail}$, max number of failure $N_{max}$

\Ensure Trained policy $\pi$
\State Initialize $\pi$ and set $L=0$, $N_{fail} = 0$, $D=\{\}$
\While{$L \le 4$}
\State Generate augmented data $\mathbf{aug}_{L}(D_h)$
\State Append into training set $D \gets D + \mathbf{aug}_{L}(D_h)$
\State Train $\pi$ on $D$
\State Eval success rate $r_{succ} = \mathbf{eval}_{L}(\pi)$
\If{$\squad$ $r_{succ} \ge r_{up} \squad \mathbf{or} \squad N_{fail} \ge N_{max}$} 
    \State $L = L + 1$, $N_{fail} = 0$
\Else
    \State Generate more data $D \gets D + \mathbf{aug}_{L}(D_h)$
    \State $N_{fail} = N_{fail} + 1$
\EndIf
\EndWhile
\end{algorithmic}
\end{algorithm}

To address this, we propose \textbf{Sensitivity-Aware Kinematics Augmentation} to randomize object poses for human demonstrations. Instead of appending a new trajectory ahead of the original one, this method amends the action for each step in the original demo to accommodate the change in object pose  $T_{W}^{O_{new}} (T_{W}^{O_{old}})^{-1}$. 
The method includes two steps: (i) Divide the entire trajectory into several segments and compute the sensitivity of each segment; (ii) Modify the end effector pose trajectory based on the sensitivity to compute the new action.

    (i) \textbf{Sensitivity Analysis for Trajectory Segments.} Sensitivity pertains to the robustness against action noise. For example, a pre-grasp state, when the hand is close to the object, has higher sensitivity compared to a state where the hand is far away. The critical insight is that it is simpler to modify the action of those states with lower sensitivity to handle the object pose variation $\Delta T = T_{W}^{O_{new}} (T_{W}^{O_{old}})^{-1}$. The robustness (the multiplicative inverse of sensitivity) of a trajectory segment $\psi$ can be mathematically defined as follows:
\begin{equation}
\begin{split}   
    \psi_{seg} &= \exp({\max} {\delta_a}) \quad  \mathrm{s.t.} \quad \mathbf{eval}(\tau') = 1 \\
    \quad \tau' &= \{a_1, a_2, ..., a_n', ..., a_{n+K-1}', ..., a_N\} \\
    \forall i \in {seg} \quad a_i' &= a_i + \delta_a \epsilon_i, \quad \epsilon_i \sim \mathcal{N}(0, 1) \\
\end{split}
\end{equation}
    In this equation, we divide the original action trajectory $\tau$ with length $N$ into $M$ segments, each segment with size $K=N/M$. Then we perturb the action within a segment $seg$ by adding Gaussian noise of scale $\delta_a$ to the original action $\{a_{m}, a_{n+1}, ..., a_{n+k-1} \}$ while keeping all the actions outside of this segment unchanged to generate perturbed trajectory $\tau'$. We assume the action space is already normalized to $[-1, 1]$ and $\mathbf{eval}$ is a binary function indicating whether an action trajectory can successfully solve the task. Intuitively, a demonstration segment is more sensitive if a smaller perturbation can cause it to fail. This sensitivity guides us on how to adjust the action to handle a new object pose. In practice, we incrementally escalate the noise scale $\delta_a$ applied to the original action trajectory until the task fails to determine $\max {\delta_a}$

    (ii) \textbf{New End Effector Pose Trajectory.} 
    To accommodate the new object pose, the total pose change of the end effector should be the same as the change in the object pose $\Delta T$. Each action contributes a small part to this change. We distribute this "task" to each step based on sensitivity:
    \begin{equation}
    \begin{split}
    \overline{\psi}_{seg_j} &= \frac{\psi_{seg_j}}{\sum_{j=1}^{M} \psi_{seg_j}}, \quad \forall {seg_j} \\
    \Delta T_j &= exp(\overline{\psi}_{seg_j} \log (\Delta T)/K) \\
    a_{i}^{new} &= a_{i} f_i(\Delta T_{j})   
    \end{split}
    \end{equation}
    In this equation, $\overline{\psi}_{seg_j}$ is the normalized robustness, $\Delta T_j$ represents the pose modification for each step, with all states in the same segment being responsible for the same amount of modification "task" to compute new action $a_{i}^{new}$. $f_i$ is a similarity transformation in $SE(3)$ space that converts the motion from the world frame to the current end effector frame. Intuitively, segments with higher robustness are tasked with more significant changes.

Please note that all the actions discussed above pertain solely to the 6D delta pose of the end effector and do not include the finger motion of the dexterous hand. For tasks such as pick-and-place or pouring, which also involve a target pose (e.g., the plate pose in pick-and-place or the bowl pose in pouring), we can apply the same augmentation strategy to the target pose (as illustrated in Level 3 of Fig.~\ref{fig:pipeline}).

\subsection{Learning Sim2Real Policy}
\label{sec:learning}

Given an augmented simulation dataset, we train a visual manipulation policy that takes images and robot proprioception as input to predict the robot's actions. In human teleoperation demonstrations, robot movements are neither Moravian nor temporally correlated. To deal with this issue, our policy is trained to predict action chunks rather than per-step actions, using Action Chunking with Transformers(ACT)~\cite{aloha}. This approach produces smoother trajectories and reduces compounding errors.

Despite our data augmentation's capacity to accommodate diverse visual and dynamic conditions, a sim2real gap remains for the robot controller. This gap becomes more challenging in our tasks, where the end effector is a high-DoF multi-finger dexterous hand. This controller gap can significantly impact non-quasi-static tasks like rotating a valve, as shown in the second row of Figure~\ref{fig:teaser}.
To close this gap, we fine-tune our network using a small set of real-world demonstrations (3-minute trajectory). However, due to the discrepancies in data collection patterns of human demos between simulation and reality, direct fine-tuning on real data risks overfitting. To ensure a smoother sim2real transfer, we employ several techniques, which will be discussed subsequently.

\noindent
\textbf{Automatic Curriculum Learning.}
Curriculum learning and data augmentation techniques are often used together to provide a smoother training process. Following the spirit of curriculum design in previous reinforcement learning work~\cite{akkaya2019solving, dextreme}, we devise a curriculum learning strategy applicable to our imitation learning context. Prior to training, we group the augmentations in Section~\ref{sec:aug} into four levels of increasing complexity, as depicted in Figure~\ref{fig:pipeline}.
As per Algorithm~\ref{alg:adr}, we begin training from the simplest level, $L=0$, signifying no augmentation, and then evaluate the task success rate after several steps of training. The evaluation difficulty aligns with the current level of $L$.
When the success rate surpasses a pre-defined threshold, we advance to the next level, which brings greater augmentation and harder evaluation. 
If the success rate fails to reach the threshold, we create additional augmented training data and stays at the current level. We continue this iterative process until all levels are completed. To prevent endless training, we introduce a fail-safe $N_{max}$: if the policy repeatedly fails during evaluation for $N_{max}$ times, we also progress to the next level.
This curriculum learning approach significantly depends on data augmentation techniques to generate training data dynamically with suitable levels of randomization. This concept stands in contrast to typical supervised learning scenarios, where data is pre-established prior to training. This on-demand data generation and customization highlights the advantage of simulation data over real-world demonstrations.

\noindent
\textbf{Action Aggregation for Small Motion.}
Human demonstrations often include noise, especially during operations involving a dexterous hand. For example, minor shaking and unintentional halting can occur within the demonstration trajectory, potentially undermining the training process. To solve this, we aggregate steps characterized by small motions, merging these actions into a single action. In practice, we set thresholds for both end-effector and finger motions to discern whether a given motion qualifies as small. 
Through the aggregation process, we can eliminate small operational noises from human actions, enabling the imitation learning policy to extract meaningful information from the state-action trajectory.

\section{Experiment Setups}
\label{sec:exp}

\begin{table*}[t]
\resizebox{\textwidth}{!}{%
\begin{tabular}{c|cccc|cccc|cccc|cccc}
\toprule
\multicolumn{1}{l|}{} & \multicolumn{4}{c|}{\begin{tabular}[c]{@{}c@{}}Pick and Place \\ Mustard Bottle \\ (Single Object)\end{tabular}} & \multicolumn{4}{c|}{\begin{tabular}[c]{@{}c@{}}Pick and Place \\ Tomato Soup Can \\ (Single Object)\end{tabular}} & \multicolumn{4}{c|}{Pouring} & \multicolumn{4}{c}{Rotating} \\  
\multicolumn{1}{l|}{} & Level 1 & Level 2 & Level 3 & Level 4 & Level 1 & Level 2 & Level 3 & Level 4 & Level 1 & Level 2 & Level 3 & Level 4 & Level 1 & Level 2 & Level 3 & Level 4 \\ \hline
R3M & 2 / 20& 0 / 20& 0 / 20 & 0 / 20& 7 / 20 & 3 / 20 & 4 / 20 & 0 / 20 & 3 / 20 & 0 / 20 & 0 / 20 & 0 / 20 & 11 / 20 & 2 / 20 & 6 / 20 & 2 / 20 \\
PVR & 4 / 20 & 0 / 20 & 0 / 20& 0 / 20& 4 / 20 & 0 / 20 & 3 / 20 & 0 / 20 & 2 / 20 & 0 / 20 & 1 / 20 & 0 / 20 & 8 / 20 & 3 / 20 & 5 / 20 & 1 / 20 \\
MVP & 2 / 20 & 0 / 20 & 3 / 20& 1 / 20 & 7 / 20 & 2 / 20 & 4 / 20 & 2 / 20 & 1 / 20 & 1 / 20 & 3 / 20 & 2 / 20 & 8 / 20 & 4 / 20 & 10 / 20 & 6 / 20 \\
Ours & \textbf{7 / 20} & \textbf{6 / 20} & \textbf{8 / 20} & \textbf{5 / 20}& \textbf{14 / 20} & \textbf{11 / 20 }& \textbf{13 / 20} & \textbf{13 / 20} & \textbf{9 / 20} & \textbf{4 / 20} & \textbf{10 / 20} & \textbf{7 / 20} & \textbf{15 / 20} & \textbf{10 / 20} & \textbf{17 / 20} & \textbf{13 / 20}
\end{tabular}}
\caption{\textbf{Main Comparison on Real Robot.} In our study, we compare the performance across four distinct tasks: (a) Pick and Place Bottle, (b) Pick and Place Can (exploring different grasping approaches), (c) Pouring (grasping a bottle and pouring its contents into a bowl), and (d) Rotating the tri-valve. We perform evaluations of the models in four levels of real-world scenarios. These levels included: (a) Level 1: In Domain, (b) Level 2: Out of Position, (c) Level 3: Random Light, and (d) Level 4: Out of Position and Random Light.}
\label{tab: main_results}
\vspace{-1.5em}
\end{table*}

Our experimental design aims to address the following key queries:

(i) How does simulation-based data augmentation compare to learning from real demonstrations in terms of both robustness and generalizability?

(ii) How does our automatic curriculum learning contribute to improved policy learning?

(iii) What is the ideal ratio between simulated and real data to train an effective policy for a real-world robot?

\subsection{Dexterous Manipulation Tasks}
We have designed three types of manipulation tasks in both real-world and simulated environments, including two quasi-static tasks (pick and place, pour) and one non-quasi-static task (rotate). For the experiments, we utilize an Allegro hand attached to an XArm6. The action space comprises a 6-dim delta end effector pose of the robot arm and a 16-dim finger joint position of the dexterous hand, with PD control employed for both arm and hand.

\textit{Pick and Place.} This task requires the robot to lift an object from the table and position it on a plate (first row of Figure~\ref{fig:teaser}). Success is achieved when the object is properly placed onto the red plate. We select two objects during data collection and testing on multiple different objects.

\textit{Rotate.} This task requires the robot to rotate a valve on the table (second row of Figure~\ref{fig:teaser}). The valve is constructed with a fixed base and a moving valve geometry, connected via a revolute joint. The task is successful when the robot rotates the valve to $720$ degrees. We use a tri-valve in data collection and test on tetra-valves and penta-valves.

\textit{Pour.} This task requires the robot to pour small boxes from a bottle into a bowl (third row of Figure~\ref{fig:teaser}). It involves three steps: (i) Lift the bottle; (ii) Move it close to the bowl; (iii) Rotate the bottle to dispense the small boxes into the bowl. Success is achieved when all four boxes have been poured into the bowl.

For each task, we have designed levels for both data augmentation ( Section~\ref{sec:aug}) and curriculum learning (Section~\ref{sec:learning}). More details regarding the design of task levels can be found in the supplementary material.

\subsection{Baselines}
Our approach can be interpreted as an initial pretraining phase using augmented simulation demonstrations followed by fine-tuning with a limited set of real data. It is natural to compare our method with other pre-training models for robotic manipulation. We have chosen three representative vision pre-training models. For all of them, we utilize the pre-trained model provided by the author and then fine-tune it using our real-world demonstration dataset.

\noindent
\textbf{PVR} is built on MoCo-v2~\cite{moco-v2}, using a ResNet50 backbone~\cite{he2016deep} trained on ImageNet~\cite{russakovsky2015imagenet}. 

\noindent
\textbf{MVP} employs self-supervised learning from a Masked Autoencoder~\cite{mae} to train visual representation on individual frames from an extensive human interaction dataset compiled from multiple existing datasets. MVP integrates a Vision Transformer~\cite{vit} backbone that segments frames into 16x16 patches. 

\noindent
\textbf{R3M} proposes a pre-training approach where a ResNet50 backbone is trained using a mix of time-contrastive learning, video-language alignment, and L1 regularization. This model is trained on a large-scale human interaction videos dataset from Ego4D~\cite{ego4d}.

\begin{table*}[]
\resizebox{\textwidth}{!}{%
\begin{tabular}{c|cccc|cccc}
\toprule
\multicolumn{1}{l|}{} & \multicolumn{4}{c|}{\begin{tabular}[c]{@{}c@{}}Test in Sim \end{tabular}} & \multicolumn{4}{c}{\begin{tabular}[c]{@{}c@{}}Test in Real\end{tabular}} \\
Set of Levels / Num of demos & Level 1 & Level 2 & Level 3 & Level 4 & In Domain & Random Light & Out of Position & \begin{tabular}[c]{@{}c@{}}Out of Position \\ + Random Light\end{tabular} \\ \hline
{[}1{]} / 100 & 78\% & 0\% & 0\% & 0\% & 20\% & 5\% & 0\% & 0\% \\
{[}1, 2{]} / 330 & 73\% & 75\% & 10.5\% & 7.5\% & 15\% & 25\% & 0\% & 0\%\\
{[}1,2,3{]} / 550 & 58\% & 66.5\% & 43.5\% & 21\% & 15\% & 15\% & 5\% & 15\% \\
{[}1,2,3,4{]} / 810 & 92.5\% & 81\% & 63\% & 49\% & 35\% & 30\% & 30\% & 40\%
\end{tabular}%
}
\caption{\textbf{Ablation on Data Augmentation.} To demonstrate the benefits of data augmentation, we employed auto-curriculum learning on various sets of levels. We performed 200 simulations to test its impact in a simulated environment and conducted 20 real-world tests to evaluate its effectiveness in a practical setting.}
\label{tab: ablation on data augmentation}
\vspace{-1em}
\end{table*}

\section{Results}
\label{sec:result}

\subsection{Main Comparison}
\textbf{Augmented simulation data markedly boosts real-world dexterous manipulation.} As depicted in Table~\ref{tab: main_results}, our methodology outperforms the baselines trained exclusively on real data in the in-domain setting (Level 1), exhibiting an average performance boost of $31.67\%$ averaged on all tasks.
Additionally, in other settings like random lighting (Level 2), out of position (Level 3), combined random lighting and out of position (Level 3) R3M, and MVP, the baselines exhibit a significant drop in success rates. In contrast, our method shows resilience to these variations, underscoring the efficacy of simulation data augmentation. Not only does this approach bridge the sim2real gap and amplify performance in in-domain real-world tasks, but it also significantly improves manipulations in out-of-domain real-world scenarios. By integrating augmentation for both visual and dynamic variations, our method successfully navigates challenges and delivers impressive results.

\subsection{Generalization to Novel Objects}
\begin{figure}[t]
  \centering
  \includegraphics[width=1\linewidth]{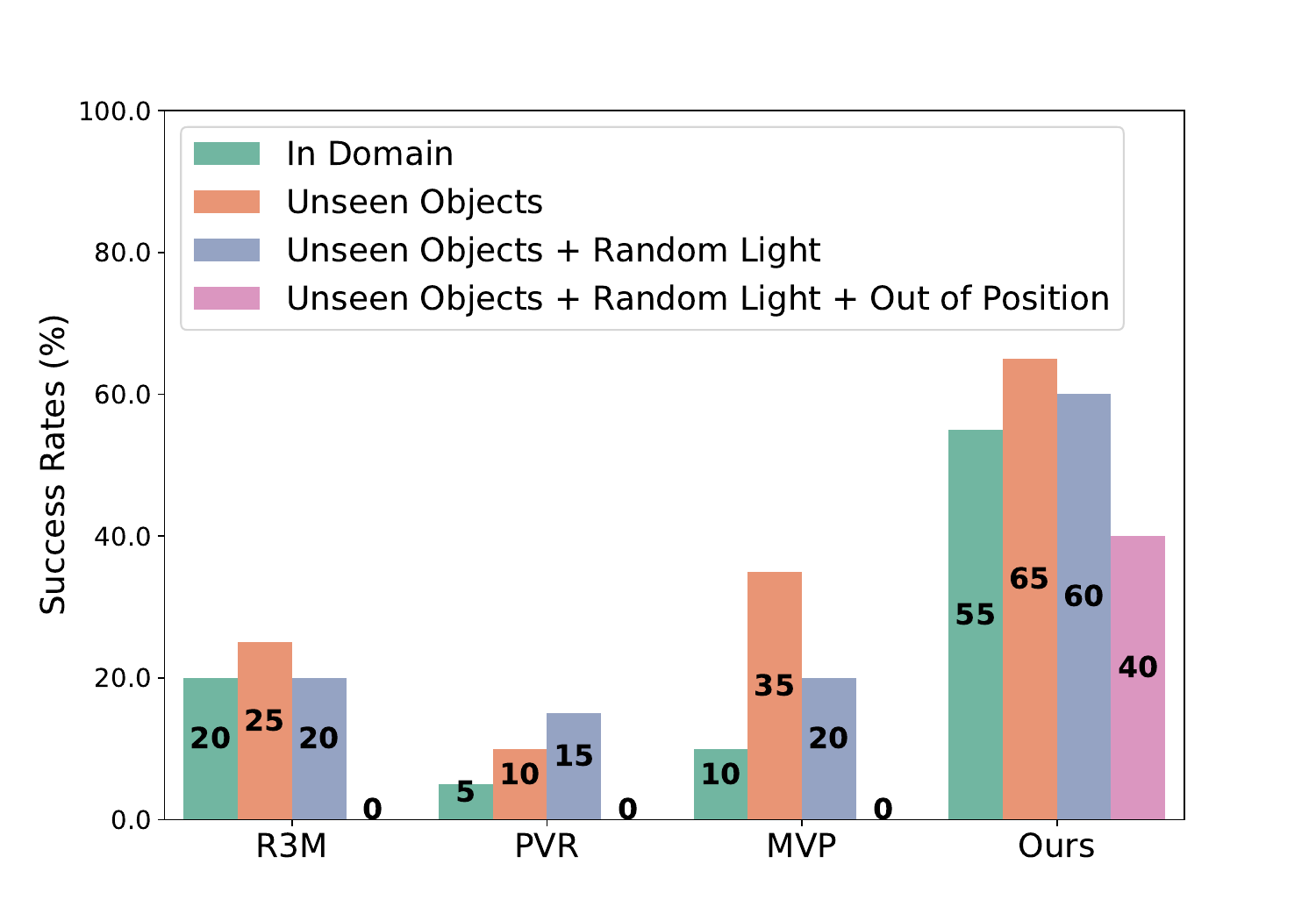}
   \caption{\textbf{Generalization to Novel Objects for Pick and Place.} We compare our approach with the baselines in scenarios involving novel objects, random light disturbances, and random object positions.}
   \label{fig:pick_place_diverse_objects}
   \vspace{-1.4em}
\end{figure}

\begin{figure}[t]
  \centering
  \includegraphics[width=1\linewidth]{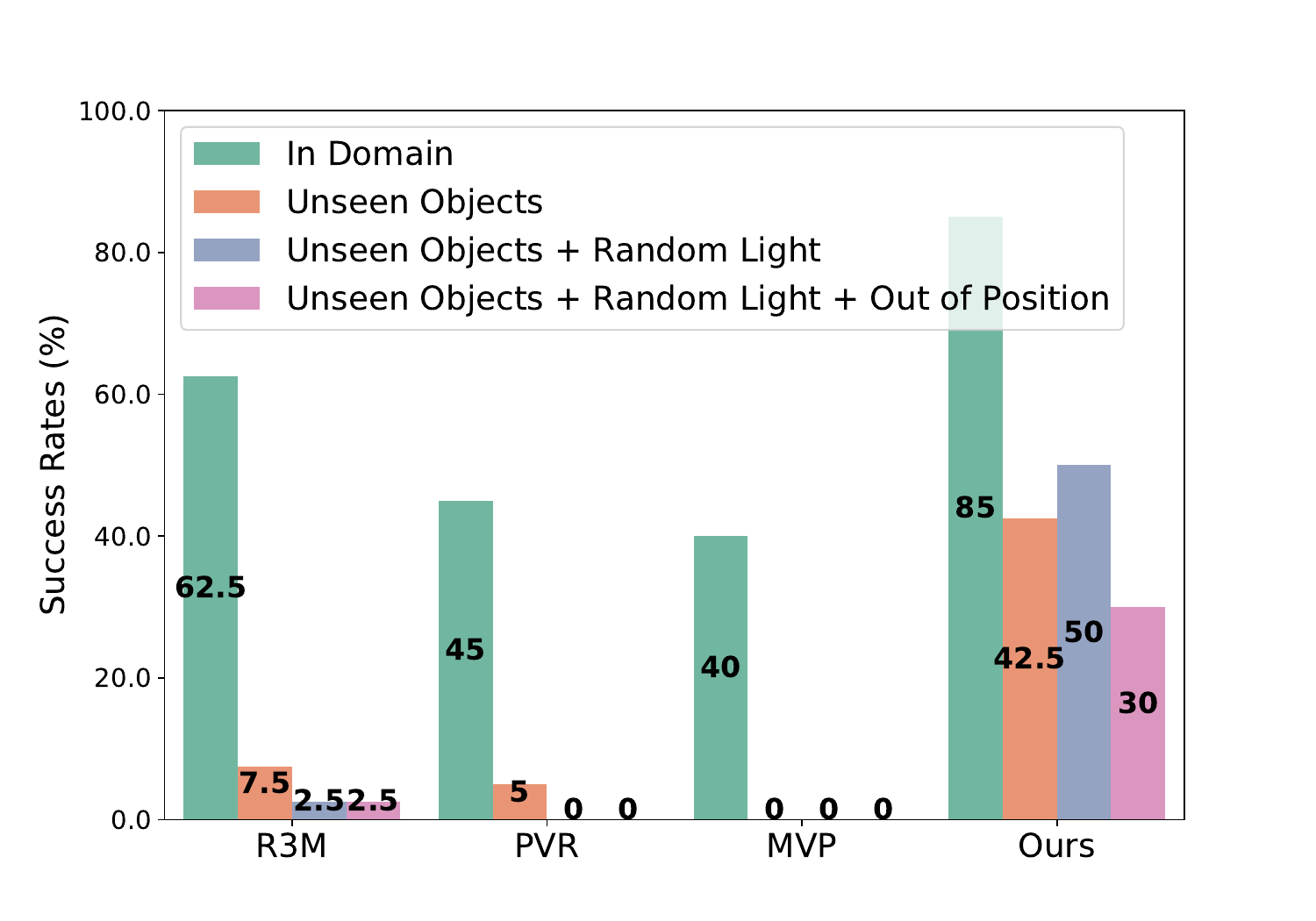}
   \caption{\textbf{Generalization to Novel Objects for Rotating.} The experimental setup for this task mirrors that of the "Generalization to Novel Objects for Pick and Place" experiments.}
   \label{fig:rotating_diverse_objects}
   \vspace{-1.4em}
\end{figure}

By incorporating data augmentation techniques, such as including diverse objects in simulation, our model can effectively manipulate unfamiliar objects, even when transitioning to a real-world context.
As shown in Figure ~\ref{fig:pick_place_diverse_objects} and ~\ref{fig:rotating_diverse_objects}, the baseline methods grapple with more complex real-world situations. In the most challenging scenario, rotating novel objects under random light conditions and new object positions, only one baseline method manages to solve it by chance with a $2.5\%$ success rate. In contrast, our method still accomplishes the task with a success rate of $30\%$.


\subsection{Ablation on Data Augmentation}
To evaluate the effectiveness of the data augmentation techniques, we perform an ablation study where our policy is trained with four levels of augmentation. As depicted in Table~\ref{tab: ablation on data augmentation}, the policy performs better in both the simulation and real-world settings with increased data augmentation, and the policy trained on all four levels excels in all metrics. Interestingly, the policy manages to solve simpler settings more effectively even in simulation when more randomness is introduced in the training data. These experiments underscore the importance of simulator-based data augmentation.

\subsection{Ablation on Auto-Curriculum Learning}
In this experiment, we evaluate the policy's effectiveness by testing it 200 times in simulations and conducting 20 real-world tests.
As shown in Table~\ref{tab:auto-curriculum}, employing curriculum learning with auto-domain randomization solely based on the data generation rate yields inferior results compared to the approach based on model performance.

\begin{table}[]
\resizebox{\columnwidth}{!}{%
\begin{tabular}{c|cccc|clll}
\toprule
\multicolumn{1}{l|}{} & \multicolumn{4}{c|}{\multirow{2}{*}{Test in Sim}} & \multicolumn{4}{c}{\multirow{2}{*}{Test in Real}} \\
\multicolumn{1}{l|}{} & \multicolumn{4}{c|}{} & \multicolumn{4}{c}{} \\
Method & Level 1 & Level 2 & Level 3 & Level 4 & \multicolumn{4}{c}{\begin{tabular}[c]{@{}c@{}}Out of Position \\ + Random Light\end{tabular}} \\ \hline
ACL (Task) & \textbf{80\%} &\textbf{61\%} & 43.5\% & 57\% & \multicolumn{4}{c}{\textbf{35\%}} \\
ACL (Data) & 19.5\% & 30\% & \textbf{75\%} & \textbf{66\%} & \multicolumn{4}{c}{20\%} \\
ACL wo CL(Data) & 20\% & 22\% & 32.5\% & 15\% & \multicolumn{4}{c}{5\%}
\end{tabular}%
}
\caption{\textbf{Ablation on Auto-Curriculum Learning}. We compare three different settings: (1) Auto Curriculum Learning based on the success rate. (2) Auto Curriculum Learning based on Data Generation Rate(the ratio of successfully generated trajectories to the total number of attempts). (3) Automatic Domain Randomization only based on Data Generation Rate. }
\label{tab:auto-curriculum}
\end{table}

\subsection{Ablation on Ratio of Sim and Real Demos}
To determine the optimal ratio between sim and real demos, we conducted tests using different combinations of sim and real demonstrations, as shown in Table ~\ref{tab:Ablation on Ratio of sim and real demos}. 
We observe that training solely on 50 real demonstrations results in poor performance, and the policy overfits to joint positions rather than utilizing the visual information in images. The best results were obtained with a combination of 15 simulation demonstrations and 35 real demonstrations. These results highlight that collecting simulation data can be exceptionally valuable, even more so considering the significantly lower data collection costs.

\begin{table}[]
\resizebox{\columnwidth}{!}{%
\begin{tabular}{c|cccc|cc}
\toprule
\multicolumn{1}{l|}{} & \multicolumn{4}{c|}{\multirow{2}{*}{Test in Sim}} & \multicolumn{2}{c}{\multirow{2}{*}{Test in Real}} \\
\multicolumn{1}{l|}{} & \multicolumn{4}{c|}{} & \multicolumn{2}{c}{} \\
Dataset & Level 1 & Level 2 & Level 3 & Level 4 & In Domain & \multicolumn{1}{l}{Out of Position} \\ \hline
50 sim & 73.5\% & \textbf{80\%} & \textbf{63.5\%} & 36\% & 0\% & 0\% \\
35 sim + 15 real & \textbf{77\%} & 70.5\% & 61\% & \textbf{45.5\%} & 25\% & \textbf{35\%} \\
15 sim + 35 real & 63\% & 74\% & 55\% & 33.5\% & \textbf{50\%} & 15\% \\
50 real & 0\% & 0\% & 0\% & 0\% & 10\% & 0\%
\end{tabular}%
}
\caption{\textbf{Ablation on Ratio of Sim and Real Demos.} We compare the performance resulting from various quantities of simulated and real demonstrations, keeping the total number of demonstrations constant.}
\label{tab:Ablation on Ratio of sim and real demos}
\vspace{-1.2em}
\end{table}

\section{Discussion}
\label{sec:discussion}

We propose CyberDemo, a novel pipeline for imitation learning in robotic manipulation, leveraging demonstrations collected in simulation. While the common belief suggests that real-world demonstrations are the optimal way to solve real-world problems, we challenge this notion by demonstrating that extensive data augmentation can make simulation data even more valuable than real-world demonstrations, a fact also supported by our experiments.
One limitation is the necessity to design a simulated environment for each real-world task, thereby increasing the human effort involved. However, since our method doesn't demand the design of specific rewards as in reinforcement learning tasks, which is often the most challenging aspect, the overall effort required is not as significant.

\section{Acknowledgement}
We express our gratitude to Haichuan Che and Chengzhe Jia for their insightful conversations regarding the repair of the Allegro hand hardware. We are also thankful to Nicklas Hansen and Jiarui Xu for their valuable input on visual pretraining. Special thanks are due to Tony Zhao for his generous contribution to the community through his open-source projects on Action Chunking with Transformers.

{
    \small
    \bibliographystyle{ieeenat_fullname}
    \bibliography{main}

\begin{thebibliography}{86}
\providecommand{\natexlab}[1]{#1}
\providecommand{\url}[1]{\texttt{#1}}
\expandafter\ifx\csname urlstyle\endcsname\relax
  \providecommand{\doi}[1]{doi: #1}\else
  \providecommand{\doi}{doi: \begingroup \urlstyle{rm}\Url}\fi

\bibitem[Akkaya et~al.(2019)Akkaya, Andrychowicz, Chociej, Litwin, McGrew, Petron, Paino, Plappert, Powell, Ribas, et~al.]{akkaya2019solving}
Ilge Akkaya, Marcin Andrychowicz, Maciek Chociej, Mateusz Litwin, Bob McGrew, Arthur Petron, Alex Paino, Matthias Plappert, Glenn Powell, Raphael Ribas, et~al.
\newblock Solving rubik's cube with a robot hand.
\newblock \emph{arXiv preprint arXiv:1910.07113}, 2019.

\bibitem[Arunachalam et~al.(2023{\natexlab{a}})Arunachalam, G{\"u}zey, Chintala, and Pinto]{arunachalam2023holo}
Sridhar~Pandian Arunachalam, Irmak G{\"u}zey, Soumith Chintala, and Lerrel Pinto.
\newblock Holo-dex: Teaching dexterity with immersive mixed reality.
\newblock In \emph{2023 IEEE International Conference on Robotics and Automation (ICRA)}, pages 5962--5969. IEEE, 2023{\natexlab{a}}.

\bibitem[Arunachalam et~al.(2023{\natexlab{b}})Arunachalam, Silwal, Evans, and Pinto]{arunachalam2023dexterous}
Sridhar~Pandian Arunachalam, Sneha Silwal, Ben Evans, and Lerrel Pinto.
\newblock Dexterous imitation made easy: A learning-based framework for efficient dexterous manipulation.
\newblock In \emph{2023 ieee international conference on robotics and automation (icra)}, pages 5954--5961. IEEE, 2023{\natexlab{b}}.

\bibitem[Aytar et~al.(2018)Aytar, Pfaff, Budden, Paine, Wang, and De~Freitas]{aytar2018playing}
Yusuf Aytar, Tobias Pfaff, David Budden, Thomas Paine, Ziyu Wang, and Nando De~Freitas.
\newblock Playing hard exploration games by watching youtube.
\newblock \emph{Advances in neural information processing systems}, 31, 2018.

\bibitem[Bharadhwaj et~al.(2023{\natexlab{a}})Bharadhwaj, Gupta, and Tulsiani]{bharadhwaj2023visual}
Homanga Bharadhwaj, Abhinav Gupta, and Shubham Tulsiani.
\newblock Visual affordance prediction for guiding robot exploration.
\newblock \emph{arXiv preprint arXiv:2305.17783}, 2023{\natexlab{a}}.

\bibitem[Bharadhwaj et~al.(2023{\natexlab{b}})Bharadhwaj, Vakil, Sharma, Gupta, Tulsiani, and Kumar]{roboagent}
Homanga Bharadhwaj, Jay Vakil, Mohit Sharma, Abhinav Gupta, Shubham Tulsiani, and Vikash Kumar.
\newblock Roboagent: Generalization and efficiency in robot manipulation via semantic augmentations and action chunking.
\newblock \emph{arXiv preprint arXiv:2309.01918}, 2023{\natexlab{b}}.

\bibitem[Black et~al.(2023)Black, Nakamoto, Atreya, Walke, Finn, Kumar, and Levine]{black2023zero}
Kevin Black, Mitsuhiko Nakamoto, Pranav Atreya, Homer Walke, Chelsea Finn, Aviral Kumar, and Sergey Levine.
\newblock Zero-shot robotic manipulation with pretrained image-editing diffusion models.
\newblock \emph{arXiv preprint arXiv:2310.10639}, 2023.

\bibitem[Bousmalis et~al.(2017)Bousmalis, Silberman, Dohan, Erhan, and Krishnan]{bousmalis2017unsupervised}
Konstantinos Bousmalis, Nathan Silberman, David Dohan, Dumitru Erhan, and Dilip Krishnan.
\newblock Unsupervised pixel-level domain adaptation with generative adversarial networks.
\newblock In \emph{Proceedings of the IEEE conference on computer vision and pattern recognition}, pages 3722--3731, 2017.

\bibitem[Brohan et~al.(2022)Brohan, Brown, Carbajal, Chebotar, Dabis, Finn, Gopalakrishnan, Hausman, Herzog, Hsu, et~al.]{rt-1}
Anthony Brohan, Noah Brown, Justice Carbajal, Yevgen Chebotar, Joseph Dabis, Chelsea Finn, Keerthana Gopalakrishnan, Karol Hausman, Alex Herzog, Jasmine Hsu, et~al.
\newblock Rt-1: Robotics transformer for real-world control at scale.
\newblock \emph{arXiv preprint arXiv:2212.06817}, 2022.

\bibitem[Caron et~al.(2021)Caron, Touvron, Misra, J{\'e}gou, Mairal, Bojanowski, and Joulin]{dino}
Mathilde Caron, Hugo Touvron, Ishan Misra, Herv{\'e} J{\'e}gou, Julien Mairal, Piotr Bojanowski, and Armand Joulin.
\newblock Emerging properties in self-supervised vision transformers.
\newblock In \emph{Proceedings of the IEEE/CVF international conference on computer vision}, pages 9650--9660, 2021.

\bibitem[Chebotar et~al.(2019)Chebotar, Handa, Makoviychuk, Macklin, Issac, Ratliff, and Fox]{chebotar2019closing}
Yevgen Chebotar, Ankur Handa, Viktor Makoviychuk, Miles Macklin, Jan Issac, Nathan Ratliff, and Dieter Fox.
\newblock Closing the sim-to-real loop: Adapting simulation randomization with real world experience.
\newblock In \emph{2019 International Conference on Robotics and Automation (ICRA)}, pages 8973--8979. IEEE, 2019.

\bibitem[Chen et~al.(2023{\natexlab{a}})Chen, Tippur, Wu, Kumar, Adelson, and Agrawal]{chen2023visual}
Tao Chen, Megha Tippur, Siyang Wu, Vikash Kumar, Edward Adelson, and Pulkit Agrawal.
\newblock Visual dexterity: In-hand reorientation of novel and complex object shapes.
\newblock \emph{Science Robotics}, 8\penalty0 (84):\penalty0 eadc9244, 2023{\natexlab{a}}.

\bibitem[Chen et~al.(2020)Chen, Fan, Girshick, and He]{moco-v2}
Xinlei Chen, Haoqi Fan, Ross Girshick, and Kaiming He.
\newblock Improved baselines with momentum contrastive learning.
\newblock \emph{arXiv preprint arXiv:2003.04297}, 2020.

\bibitem[Chen et~al.(2022)Chen, Van~Wyk, Chao, Yang, Mousavian, Gupta, and Fox]{chen2022learning}
Zoey Chen, Karl Van~Wyk, Yu-Wei Chao, Wei Yang, Arsalan Mousavian, Abhishek Gupta, and Dieter Fox.
\newblock Learning robust real-world dexterous grasping policies via implicit shape augmentation.
\newblock \emph{arXiv preprint arXiv:2210.13638}, 2022.

\bibitem[Chen et~al.(2023{\natexlab{b}})Chen, Kiami, Gupta, and Kumar]{genaug}
Zoey Chen, Sho Kiami, Abhishek Gupta, and Vikash Kumar.
\newblock Genaug: Retargeting behaviors to unseen situations via generative augmentation.
\newblock \emph{arXiv preprint arXiv:2302.06671}, 2023{\natexlab{b}}.

\bibitem[Cubuk et~al.(1805)Cubuk, Zoph, Mane, Vasudevan, and Le]{cubuk1805autoaugment}
Ekin~D Cubuk, Barret Zoph, Dandelion Mane, Vijay Vasudevan, and Quoc~V Le.
\newblock Autoaugment: Learning augmentation policies from data. arxiv 2018.
\newblock \emph{arXiv preprint arXiv:1805.09501}, 1805.

\bibitem[Dalal et~al.(2023)Dalal, Mandlekar, Garrett, Handa, Salakhutdinov, and Fox]{dalal2023imitating}
Murtaza Dalal, Ajay Mandlekar, Caelan Garrett, Ankur Handa, Ruslan Salakhutdinov, and Dieter Fox.
\newblock Imitating task and motion planning with visuomotor transformers.
\newblock \emph{arXiv preprint arXiv:2305.16309}, 2023.

\bibitem[Deng et~al.(2009)Deng, Dong, Socher, Li, Li, and Fei-Fei]{imagenet}
Jia Deng, Wei Dong, Richard Socher, Li-Jia Li, Kai Li, and Li Fei-Fei.
\newblock Imagenet: A large-scale hierarchical image database.
\newblock In \emph{2009 IEEE conference on computer vision and pattern recognition}, pages 248--255. Ieee, 2009.

\bibitem[Dosovitskiy et~al.(2020)Dosovitskiy, Beyer, Kolesnikov, Weissenborn, Zhai, Unterthiner, Dehghani, Minderer, Heigold, Gelly, et~al.]{vit}
Alexey Dosovitskiy, Lucas Beyer, Alexander Kolesnikov, Dirk Weissenborn, Xiaohua Zhai, Thomas Unterthiner, Mostafa Dehghani, Matthias Minderer, Georg Heigold, Sylvain Gelly, et~al.
\newblock An image is worth 16x16 words: Transformers for image recognition at scale.
\newblock \emph{arXiv preprint arXiv:2010.11929}, 2020.

\bibitem[Ebert et~al.(2021)Ebert, Yang, Schmeckpeper, Bucher, Georgakis, Daniilidis, Finn, and Levine]{bridge-data}
Frederik Ebert, Yanlai Yang, Karl Schmeckpeper, Bernadette Bucher, Georgios Georgakis, Kostas Daniilidis, Chelsea Finn, and Sergey Levine.
\newblock Bridge data: Boosting generalization of robotic skills with cross-domain datasets.
\newblock \emph{arXiv preprint arXiv:2109.13396}, 2021.

\bibitem[Ganin et~al.(2016)Ganin, Ustinova, Ajakan, Germain, Larochelle, Laviolette, Marchand, and Lempitsky]{dann}
Yaroslav Ganin, Evgeniya Ustinova, Hana Ajakan, Pascal Germain, Hugo Larochelle, Fran{\c{c}}ois Laviolette, Mario Marchand, and Victor Lempitsky.
\newblock Domain-adversarial training of neural networks.
\newblock \emph{The journal of machine learning research}, 17\penalty0 (1):\penalty0 2096--2030, 2016.

\bibitem[Grauman et~al.(2022)Grauman, Westbury, Byrne, Chavis, Furnari, Girdhar, Hamburger, Jiang, Liu, Liu, et~al.]{ego4d}
Kristen Grauman, Andrew Westbury, Eugene Byrne, Zachary Chavis, Antonino Furnari, Rohit Girdhar, Jackson Hamburger, Hao Jiang, Miao Liu, Xingyu Liu, et~al.
\newblock Ego4d: Around the world in 3,000 hours of egocentric video.
\newblock In \emph{Proceedings of the IEEE/CVF Conference on Computer Vision and Pattern Recognition}, pages 18995--19012, 2022.

\bibitem[Gu et~al.(2023)Gu, Xiang, Li, Ling, Liu, Mu, Tang, Tao, Wei, Yao, et~al.]{maniskill2}
Jiayuan Gu, Fanbo Xiang, Xuanlin Li, Zhan Ling, Xiqiang Liu, Tongzhou Mu, Yihe Tang, Stone Tao, Xinyue Wei, Yunchao Yao, et~al.
\newblock Maniskill2: A unified benchmark for generalizable manipulation skills.
\newblock \emph{arXiv preprint arXiv:2302.04659}, 2023.

\bibitem[Handa et~al.(2023)Handa, Allshire, Makoviychuk, Petrenko, Singh, Liu, Makoviichuk, Van~Wyk, Zhurkevich, Sundaralingam, et~al.]{dextreme}
Ankur Handa, Arthur Allshire, Viktor Makoviychuk, Aleksei Petrenko, Ritvik Singh, Jingzhou Liu, Denys Makoviichuk, Karl Van~Wyk, Alexander Zhurkevich, Balakumar Sundaralingam, et~al.
\newblock Dextreme: Transfer of agile in-hand manipulation from simulation to reality.
\newblock In \emph{2023 IEEE International Conference on Robotics and Automation (ICRA)}, pages 5977--5984. IEEE, 2023.

\bibitem[He et~al.(2016)He, Zhang, Ren, and Sun]{he2016deep}
Kaiming He, Xiangyu Zhang, Shaoqing Ren, and Jian Sun.
\newblock Deep residual learning for image recognition.
\newblock In \emph{Proceedings of the IEEE conference on computer vision and pattern recognition}, pages 770--778, 2016.

\bibitem[He et~al.(2020)He, Fan, Wu, Xie, and Girshick]{moco}
Kaiming He, Haoqi Fan, Yuxin Wu, Saining Xie, and Ross Girshick.
\newblock Momentum contrast for unsupervised visual representation learning.
\newblock In \emph{Proceedings of the IEEE/CVF conference on computer vision and pattern recognition}, pages 9729--9738, 2020.

\bibitem[He et~al.(2022)He, Chen, Xie, Li, Doll{\'a}r, and Girshick]{mae}
Kaiming He, Xinlei Chen, Saining Xie, Yanghao Li, Piotr Doll{\'a}r, and Ross Girshick.
\newblock Masked autoencoders are scalable vision learners.
\newblock In \emph{Proceedings of the IEEE/CVF conference on computer vision and pattern recognition}, pages 16000--16009, 2022.

\bibitem[Ho et~al.(2021)Ho, Rao, Xu, Jang, Khansari, and Bai]{retinagan}
Daniel Ho, Kanishka Rao, Zhuo Xu, Eric Jang, Mohi Khansari, and Yunfei Bai.
\newblock Retinagan: An object-aware approach to sim-to-real transfer.
\newblock In \emph{2021 IEEE International Conference on Robotics and Automation (ICRA)}, pages 10920--10926. IEEE, 2021.

\bibitem[Huang et~al.(2023{\natexlab{a}})Huang, Chen, Wang, Qin, Yang, Atanasov, and Wang]{huang2023dynamic}
Binghao Huang, Yuanpei Chen, Tianyu Wang, Yuzhe Qin, Yaodong Yang, Nikolay Atanasov, and Xiaolong Wang.
\newblock Dynamic handover: Throw and catch with bimanual hands.
\newblock \emph{arXiv preprint arXiv:2309.05655}, 2023{\natexlab{a}}.

\bibitem[Huang et~al.(2023{\natexlab{b}})Huang, Wang, Zhang, Li, Wu, and Fei-Fei]{huang2023voxposer}
Wenlong Huang, Chen Wang, Ruohan Zhang, Yunzhu Li, Jiajun Wu, and Li Fei-Fei.
\newblock Voxposer: Composable 3d value maps for robotic manipulation with language models.
\newblock \emph{arXiv preprint arXiv:2307.05973}, 2023{\natexlab{b}}.

\bibitem[James et~al.(2020)James, Ma, Arrojo, and Davison]{rlbench}
Stephen James, Zicong Ma, David~Rovick Arrojo, and Andrew~J Davison.
\newblock Rlbench: The robot learning benchmark \& learning environment.
\newblock \emph{IEEE Robotics and Automation Letters}, 5\penalty0 (2):\penalty0 3019--3026, 2020.

\bibitem[Jang et~al.(2022)Jang, Irpan, Khansari, Kappler, Ebert, Lynch, Levine, and Finn]{bc-z}
Eric Jang, Alex Irpan, Mohi Khansari, Daniel Kappler, Frederik Ebert, Corey Lynch, Sergey Levine, and Chelsea Finn.
\newblock Bc-z: Zero-shot task generalization with robotic imitation learning.
\newblock In \emph{Conference on Robot Learning}, pages 991--1002. PMLR, 2022.

\bibitem[Jiang et~al.(2022)Jiang, Gupta, Zhang, Wang, Dou, Chen, Fei-Fei, Anandkumar, Zhu, and Fan]{vima}
Yunfan Jiang, Agrim Gupta, Zichen Zhang, Guanzhi Wang, Yongqiang Dou, Yanjun Chen, Li Fei-Fei, Anima Anandkumar, Yuke Zhu, and Linxi Fan.
\newblock Vima: General robot manipulation with multimodal prompts.
\newblock \emph{arXiv}, 2022.

\bibitem[Kar et~al.(2019)Kar, Prakash, Liu, Cameracci, Yuan, Rusiniak, Acuna, Torralba, and Fidler]{kar2019meta}
Amlan Kar, Aayush Prakash, Ming-Yu Liu, Eric Cameracci, Justin Yuan, Matt Rusiniak, David Acuna, Antonio Torralba, and Sanja Fidler.
\newblock Meta-sim: Learning to generate synthetic datasets.
\newblock In \emph{Proceedings of the IEEE/CVF International Conference on Computer Vision}, pages 4551--4560, 2019.

\bibitem[Kaspar et~al.(2020)Kaspar, Osorio, and Bock]{kaspar2020sim2real}
Manuel Kaspar, Juan D~Mu{\~n}oz Osorio, and J{\"u}rgen Bock.
\newblock Sim2real transfer for reinforcement learning without dynamics randomization.
\newblock In \emph{2020 IEEE/RSJ International Conference on Intelligent Robots and Systems (IROS)}, pages 4383--4388. IEEE, 2020.

\bibitem[Kumar et~al.(2023)Kumar, Zamora, Hansen, Jangir, and Wang]{kumar2023graph}
Sateesh Kumar, Jonathan Zamora, Nicklas Hansen, Rishabh Jangir, and Xiaolong Wang.
\newblock Graph inverse reinforcement learning from diverse videos.
\newblock In \emph{Conference on Robot Learning}, pages 55--66. PMLR, 2023.

\bibitem[Li et~al.(2023)Li, Zhang, Wong, Gokmen, Srivastava, Mart{\'\i}n-Mart{\'\i}n, Wang, Levine, Lingelbach, Sun, et~al.]{behavior1k}
Chengshu Li, Ruohan Zhang, Josiah Wong, Cem Gokmen, Sanjana Srivastava, Roberto Mart{\'\i}n-Mart{\'\i}n, Chen Wang, Gabrael Levine, Michael Lingelbach, Jiankai Sun, et~al.
\newblock Behavior-1k: A benchmark for embodied ai with 1,000 everyday activities and realistic simulation.
\newblock In \emph{Conference on Robot Learning}, pages 80--93. PMLR, 2023.

\bibitem[Liang et~al.(2020)Liang, Saxena, and Kroemer]{liang2020learning}
Jacky Liang, Saumya Saxena, and Oliver Kroemer.
\newblock Learning active task-oriented exploration policies for bridging the sim-to-real gap.
\newblock \emph{arXiv preprint arXiv:2006.01952}, 2020.

\bibitem[Lynch et~al.(2023)Lynch, Wahid, Tompson, Ding, Betker, Baruch, Armstrong, and Florence]{interactive-language}
Corey Lynch, Ayzaan Wahid, Jonathan Tompson, Tianli Ding, James Betker, Robert Baruch, Travis Armstrong, and Pete Florence.
\newblock Interactive language: Talking to robots in real time.
\newblock \emph{IEEE Robotics and Automation Letters}, 2023.

\bibitem[Mandi et~al.(2022)Mandi, Bharadhwaj, Moens, Song, Rajeswaran, and Kumar]{cacti}
Zhao Mandi, Homanga Bharadhwaj, Vincent Moens, Shuran Song, Aravind Rajeswaran, and Vikash Kumar.
\newblock Cacti: A framework for scalable multi-task multi-scene visual imitation learning.
\newblock \emph{arXiv preprint arXiv:2212.05711}, 2022.

\bibitem[Mandikal and Grauman(2022)]{dexvip}
Priyanka Mandikal and Kristen Grauman.
\newblock Dexvip: Learning dexterous grasping with human hand pose priors from video.
\newblock In \emph{Conference on Robot Learning}, pages 651--661. PMLR, 2022.

\bibitem[Mandlekar et~al.(2018)Mandlekar, Zhu, Garg, Booher, Spero, Tung, Gao, Emmons, Gupta, Orbay, et~al.]{roboturk}
Ajay Mandlekar, Yuke Zhu, Animesh Garg, Jonathan Booher, Max Spero, Albert Tung, Julian Gao, John Emmons, Anchit Gupta, Emre Orbay, et~al.
\newblock Roboturk: A crowdsourcing platform for robotic skill learning through imitation.
\newblock In \emph{Conference on Robot Learning}, pages 879--893. PMLR, 2018.

\bibitem[Mandlekar et~al.(2021)Mandlekar, Xu, Wong, Nasiriany, Wang, Kulkarni, Fei-Fei, Savarese, Zhu, and Mart{\'\i}n-Mart{\'\i}n]{robo-mimic}
Ajay Mandlekar, Danfei Xu, Josiah Wong, Soroush Nasiriany, Chen Wang, Rohun Kulkarni, Li Fei-Fei, Silvio Savarese, Yuke Zhu, and Roberto Mart{\'\i}n-Mart{\'\i}n.
\newblock What matters in learning from offline human demonstrations for robot manipulation.
\newblock \emph{arXiv preprint arXiv:2108.03298}, 2021.

\bibitem[Mandlekar et~al.(2023)Mandlekar, Nasiriany, Wen, Akinola, Narang, Fan, Zhu, and Fox]{mimicgen}
Ajay Mandlekar, Soroush Nasiriany, Bowen Wen, Iretiayo Akinola, Yashraj Narang, Linxi Fan, Yuke Zhu, and Dieter Fox.
\newblock Mimicgen: A data generation system for scalable robot learning using human demonstrations.
\newblock \emph{arXiv preprint arXiv:2310.17596}, 2023.

\bibitem[Mees et~al.(2020)Mees, Merklinger, Kalweit, and Burgard]{mees2020adversarial}
Oier Mees, Markus Merklinger, Gabriel Kalweit, and Wolfram Burgard.
\newblock Adversarial skill networks: Unsupervised robot skill learning from video.
\newblock In \emph{2020 IEEE International Conference on Robotics and Automation (ICRA)}, pages 4188--4194. IEEE, 2020.

\bibitem[Nair et~al.(2022)Nair, Rajeswaran, Kumar, Finn, and Gupta]{r3m}
Suraj Nair, Aravind Rajeswaran, Vikash Kumar, Chelsea Finn, and Abhinav Gupta.
\newblock R3m: A universal visual representation for robot manipulation.
\newblock \emph{arXiv preprint arXiv:2203.12601}, 2022.

\bibitem[Nguyen et~al.(2018)Nguyen, Kanoulas, Muratore, Caldwell, and Tsagarakis]{nguyen2018translating}
Anh Nguyen, Dimitrios Kanoulas, Luca Muratore, Darwin~G Caldwell, and Nikos~G Tsagarakis.
\newblock Translating videos to commands for robotic manipulation with deep recurrent neural networks.
\newblock In \emph{2018 IEEE International Conference on Robotics and Automation (ICRA)}, pages 3782--3788. IEEE, 2018.

\bibitem[Padalkar et~al.(2023)Padalkar, Pooley, Jain, Bewley, Herzog, Irpan, Khazatsky, Rai, Singh, Brohan, et~al.]{rt-x}
Abhishek Padalkar, Acorn Pooley, Ajinkya Jain, Alex Bewley, Alex Herzog, Alex Irpan, Alexander Khazatsky, Anant Rai, Anikait Singh, Anthony Brohan, et~al.
\newblock Open x-embodiment: Robotic learning datasets and rt-x models.
\newblock \emph{arXiv preprint arXiv:2310.08864}, 2023.

\bibitem[Parisi et~al.(2022)Parisi, Rajeswaran, Purushwalkam, and Gupta]{pvr}
Simone Parisi, Aravind Rajeswaran, Senthil Purushwalkam, and Abhinav Gupta.
\newblock The unsurprising effectiveness of pre-trained vision models for control.
\newblock In \emph{International Conference on Machine Learning}, pages 17359--17371. PMLR, 2022.

\bibitem[Peng et~al.(2018)Peng, Andrychowicz, Zaremba, and Abbeel]{peng2018sim}
Xue~Bin Peng, Marcin Andrychowicz, Wojciech Zaremba, and Pieter Abbeel.
\newblock Sim-to-real transfer of robotic control with dynamics randomization.
\newblock In \emph{2018 IEEE international conference on robotics and automation (ICRA)}, pages 3803--3810. IEEE, 2018.

\bibitem[Qi et~al.(2023{\natexlab{a}})Qi, Kumar, Calandra, Ma, and Malik]{hora}
Haozhi Qi, Ashish Kumar, Roberto Calandra, Yi Ma, and Jitendra Malik.
\newblock In-hand object rotation via rapid motor adaptation.
\newblock In \emph{Conference on Robot Learning}, pages 1722--1732. PMLR, 2023{\natexlab{a}}.

\bibitem[Qi et~al.(2023{\natexlab{b}})Qi, Yi, Suresh, Lambeta, Ma, Calandra, and Malik]{qi2023general}
Haozhi Qi, Brent Yi, Sudharshan Suresh, Mike Lambeta, Yi Ma, Roberto Calandra, and Jitendra Malik.
\newblock General in-hand object rotation with vision and touch.
\newblock In \emph{Conference on Robot Learning}, pages 2549--2564. PMLR, 2023{\natexlab{b}}.

\bibitem[Qin et~al.(2022{\natexlab{a}})Qin, Su, and Wang]{from-one-hand}
Yuzhe Qin, Hao Su, and Xiaolong Wang.
\newblock From one hand to multiple hands: Imitation learning for dexterous manipulation from single-camera teleoperation.
\newblock \emph{RA-L}, 7\penalty0 (4):\penalty0 10873--10881, 2022{\natexlab{a}}.

\bibitem[Qin et~al.(2022{\natexlab{b}})Qin, Wu, Liu, Jiang, Yang, Fu, and Wang]{dexmv}
Yuzhe Qin, Yueh-Hua Wu, Shaowei Liu, Hanwen Jiang, Ruihan Yang, Yang Fu, and Xiaolong Wang.
\newblock Dexmv: Imitation learning for dexterous manipulation from human videos.
\newblock In \emph{European Conference on Computer Vision}, pages 570--587. Springer, 2022{\natexlab{b}}.

\bibitem[Qin et~al.(2023)Qin, Yang, Huang, Van~Wyk, Su, Wang, Chao, and Fox]{anyteleop}
Yuzhe Qin, Wei Yang, Binghao Huang, Karl Van~Wyk, Hao Su, Xiaolong Wang, Yu-Wei Chao, and Dietor Fox.
\newblock Anyteleop: A general vision-based dexterous robot arm-hand teleoperation system.
\newblock \emph{arXiv preprint arXiv:2307.04577}, 2023.

\bibitem[Rao et~al.(2020)Rao, Harris, Irpan, Levine, Ibarz, and Khansari]{rl-cyclegan}
Kanishka Rao, Chris Harris, Alex Irpan, Sergey Levine, Julian Ibarz, and Mohi Khansari.
\newblock Rl-cyclegan: Reinforcement learning aware simulation-to-real.
\newblock In \emph{Proceedings of the IEEE/CVF Conference on Computer Vision and Pattern Recognition}, pages 11157--11166, 2020.

\bibitem[Russakovsky et~al.(2015)Russakovsky, Deng, Su, Krause, Satheesh, Ma, Huang, Karpathy, Khosla, Bernstein, et~al.]{russakovsky2015imagenet}
Olga Russakovsky, Jia Deng, Hao Su, Jonathan Krause, Sanjeev Satheesh, Sean Ma, Zhiheng Huang, Andrej Karpathy, Aditya Khosla, Michael Bernstein, et~al.
\newblock Imagenet large scale visual recognition challenge.
\newblock \emph{International journal of computer vision}, 115:\penalty0 211--252, 2015.

\bibitem[Savva et~al.(2019)Savva, Kadian, Maksymets, Zhao, Wijmans, Jain, Straub, Liu, Koltun, Malik, et~al.]{savva2019habitat}
Manolis Savva, Abhishek Kadian, Oleksandr Maksymets, Yili Zhao, Erik Wijmans, Bhavana Jain, Julian Straub, Jia Liu, Vladlen Koltun, Jitendra Malik, et~al.
\newblock Habitat: A platform for embodied ai research.
\newblock In \emph{Proceedings of the IEEE/CVF international conference on computer vision}, pages 9339--9347, 2019.

\bibitem[Schmeckpeper et~al.(2020)Schmeckpeper, Rybkin, Daniilidis, Levine, and Finn]{schmeckpeper2020reinforcement}
Karl Schmeckpeper, Oleh Rybkin, Kostas Daniilidis, Sergey Levine, and Chelsea Finn.
\newblock Reinforcement learning with videos: Combining offline observations with interaction.
\newblock \emph{arXiv preprint arXiv:2011.06507}, 2020.

\bibitem[Schwarzer et~al.(2021)Schwarzer, Rajkumar, Noukhovitch, Anand, Charlin, Hjelm, Bachman, and Courville]{schwarzer2021pretraining}
Max Schwarzer, Nitarshan Rajkumar, Michael Noukhovitch, Ankesh Anand, Laurent Charlin, R~Devon Hjelm, Philip Bachman, and Aaron~C Courville.
\newblock Pretraining representations for data-efficient reinforcement learning.
\newblock \emph{Advances in Neural Information Processing Systems}, 34:\penalty0 12686--12699, 2021.

\bibitem[Sermanet et~al.(2018)Sermanet, Lynch, Chebotar, Hsu, Jang, Schaal, Levine, and Brain]{tcn}
Pierre Sermanet, Corey Lynch, Yevgen Chebotar, Jasmine Hsu, Eric Jang, Stefan Schaal, Sergey Levine, and Google Brain.
\newblock Time-contrastive networks: Self-supervised learning from video.
\newblock In \emph{2018 IEEE international conference on robotics and automation (ICRA)}, pages 1134--1141. IEEE, 2018.

\bibitem[Shah and Kumar(2021)]{shah2021rrl}
Rutav Shah and Vikash Kumar.
\newblock Rrl: Resnet as representation for reinforcement learning.
\newblock \emph{arXiv preprint arXiv:2107.03380}, 2021.

\bibitem[Shao et~al.(2021)Shao, Migimatsu, Zhang, Yang, and Bohg]{concept2robot}
Lin Shao, Toki Migimatsu, Qiang Zhang, Karen Yang, and Jeannette Bohg.
\newblock Concept2robot: Learning manipulation concepts from instructions and human demonstrations.
\newblock \emph{The International Journal of Robotics Research}, 40\penalty0 (12-14):\penalty0 1419--1434, 2021.

\bibitem[Shaw et~al.(2023)Shaw, Bahl, and Pathak]{videodex}
Kenneth Shaw, Shikhar Bahl, and Deepak Pathak.
\newblock Videodex: Learning dexterity from internet videos.
\newblock In \emph{Conference on Robot Learning}, pages 654--665. PMLR, 2023.

\bibitem[Shorten and Khoshgoftaar(2019)]{shorten2019survey}
Connor Shorten and Taghi~M Khoshgoftaar.
\newblock A survey on image data augmentation for deep learning.
\newblock \emph{Journal of big data}, 6\penalty0 (1):\penalty0 1--48, 2019.

\bibitem[Shridhar et~al.(2022)Shridhar, Manuelli, and Fox]{cliport}
Mohit Shridhar, Lucas Manuelli, and Dieter Fox.
\newblock Cliport: What and where pathways for robotic manipulation.
\newblock In \emph{Conference on Robot Learning}, pages 894--906. PMLR, 2022.

\bibitem[Singh et~al.(2023)Singh, Blukis, Mousavian, Goyal, Xu, Tremblay, Fox, Thomason, and Garg]{progprompt}
Ishika Singh, Valts Blukis, Arsalan Mousavian, Ankit Goyal, Danfei Xu, Jonathan Tremblay, Dieter Fox, Jesse Thomason, and Animesh Garg.
\newblock Progprompt: Generating situated robot task plans using large language models.
\newblock In \emph{2023 IEEE International Conference on Robotics and Automation (ICRA)}, pages 11523--11530. IEEE, 2023.

\bibitem[Sohn et~al.(2015)Sohn, Lee, and Yan]{cvae}
Kihyuk Sohn, Honglak Lee, and Xinchen Yan.
\newblock Learning structured output representation using deep conditional generative models.
\newblock \emph{Advances in neural information processing systems}, 28, 2015.

\bibitem[Stepputtis et~al.(2020)Stepputtis, Campbell, Phielipp, Lee, Baral, and Ben~Amor]{stepputtis2020language}
Simon Stepputtis, Joseph Campbell, Mariano Phielipp, Stefan Lee, Chitta Baral, and Heni Ben~Amor.
\newblock Language-conditioned imitation learning for robot manipulation tasks.
\newblock \emph{Advances in Neural Information Processing Systems}, 33:\penalty0 13139--13150, 2020.

\bibitem[Stooke et~al.(2021)Stooke, Lee, Abbeel, and Laskin]{stooke2021decoupling}
Adam Stooke, Kimin Lee, Pieter Abbeel, and Michael Laskin.
\newblock Decoupling representation learning from reinforcement learning.
\newblock In \emph{International Conference on Machine Learning}, pages 9870--9879. PMLR, 2021.

\bibitem[Tobin et~al.(2017)Tobin, Fong, Ray, Schneider, Zaremba, and Abbeel]{domain-random}
Josh Tobin, Rachel Fong, Alex Ray, Jonas Schneider, Wojciech Zaremba, and Pieter Abbeel.
\newblock Domain randomization for transferring deep neural networks from simulation to the real world.
\newblock In \emph{2017 IEEE/RSJ international conference on intelligent robots and systems (IROS)}, pages 23--30. IEEE, 2017.

\bibitem[Truong et~al.(2021)Truong, Chernova, and Batra]{truong2021bi}
Joanne Truong, Sonia Chernova, and Dhruv Batra.
\newblock Bi-directional domain adaptation for sim2real transfer of embodied navigation agents.
\newblock \emph{IEEE Robotics and Automation Letters}, 6\penalty0 (2):\penalty0 2634--2641, 2021.

\bibitem[Wang et~al.(2023)Wang, Guo, Vuong, Qin, Su, and Christensen]{real2sim2real}
Luobin Wang, Runlin Guo, Quan Vuong, Yuzhe Qin, Hao Su, and Henrik Christensen.
\newblock A real2sim2real method for robust object grasping with neural surface reconstruction.
\newblock In \emph{2023 IEEE 19th International Conference on Automation Science and Engineering (CASE)}, pages 1--8. IEEE, 2023.

\bibitem[Xiang et~al.(2020)Xiang, Qin, Mo, Xia, Zhu, Liu, Liu, Jiang, Yuan, Wang, et~al.]{sapien}
Fanbo Xiang, Yuzhe Qin, Kaichun Mo, Yikuan Xia, Hao Zhu, Fangchen Liu, Minghua Liu, Hanxiao Jiang, Yifu Yuan, He Wang, et~al.
\newblock Sapien: A simulated part-based interactive environment.
\newblock In \emph{Proceedings of the IEEE/CVF Conference on Computer Vision and Pattern Recognition}, pages 11097--11107, 2020.

\bibitem[Xiao et~al.(2022)Xiao, Radosavovic, Darrell, and Malik]{mvp}
Tete Xiao, Ilija Radosavovic, Trevor Darrell, and Jitendra Malik.
\newblock Masked visual pre-training for motor control.
\newblock \emph{arXiv preprint arXiv:2203.06173}, 2022.

\bibitem[Yen-Chen et~al.(2020)Yen-Chen, Zeng, Song, Isola, and Lin]{yen2020learning}
Lin Yen-Chen, Andy Zeng, Shuran Song, Phillip Isola, and Tsung-Yi Lin.
\newblock Learning to see before learning to act: Visual pre-training for manipulation.
\newblock In \emph{2020 IEEE International Conference on Robotics and Automation (ICRA)}, pages 7286--7293. IEEE, 2020.

\bibitem[Yin et~al.(2023)Yin, Huang, Qin, Chen, and Wang]{rotating-touch}
Zhao-Heng Yin, Binghao Huang, Yuzhe Qin, Qifeng Chen, and Xiaolong Wang.
\newblock Rotating without seeing: Towards in-hand dexterity through touch.
\newblock \emph{arXiv preprint arXiv:2303.10880}, 2023.

\bibitem[Yu et~al.(2023)Yu, Xiao, Stone, Tompson, Brohan, Wang, Singh, Tan, Peralta, Ichter, et~al.]{yu2023scaling}
Tianhe Yu, Ted Xiao, Austin Stone, Jonathan Tompson, Anthony Brohan, Su Wang, Jaspiar Singh, Clayton Tan, Jodilyn Peralta, Brian Ichter, et~al.
\newblock Scaling robot learning with semantically imagined experience.
\newblock \emph{arXiv preprint arXiv:2302.11550}, 2023.

\bibitem[Yuan et~al.(2023)Yuan, Che, Qin, Huang, Yin, Lee, Wu, Lim, and Wang]{yuan2023robot}
Ying Yuan, Haichuan Che, Yuzhe Qin, Binghao Huang, Zhao-Heng Yin, Kang-Won Lee, Yi Wu, Soo-Chul Lim, and Xiaolong Wang.
\newblock Robot synesthesia: In-hand manipulation with visuotactile sensing.
\newblock \emph{arXiv preprint arXiv:2312.01853}, 2023.

\bibitem[Yuan et~al.(2022)Yuan, Xue, Yuan, Wang, Wu, Gao, and Xu]{yuan2022pre}
Zhecheng Yuan, Zhengrong Xue, Bo Yuan, Xueqian Wang, Yi Wu, Yang Gao, and Huazhe Xu.
\newblock Pre-trained image encoder for generalizable visual reinforcement learning.
\newblock \emph{Advances in Neural Information Processing Systems}, 35:\penalty0 13022--13037, 2022.

\bibitem[Zakharov et~al.(2019)Zakharov, Kehl, and Ilic]{zakharov2019deceptionnet}
Sergey Zakharov, Wadim Kehl, and Slobodan Ilic.
\newblock Deceptionnet: Network-driven domain randomization.
\newblock In \emph{Proceedings of the IEEE/CVF International Conference on Computer Vision}, pages 532--541, 2019.

\bibitem[Zakka et~al.(2022)Zakka, Zeng, Florence, Tompson, Bohg, and Dwibedi]{xirl}
Kevin Zakka, Andy Zeng, Pete Florence, Jonathan Tompson, Jeannette Bohg, and Debidatta Dwibedi.
\newblock Xirl: Cross-embodiment inverse reinforcement learning.
\newblock In \emph{Conference on Robot Learning}, pages 537--546. PMLR, 2022.

\bibitem[Ze et~al.(2023)Ze, Hansen, Chen, Jain, and Wang]{ze2023visual}
Yanjie Ze, Nicklas Hansen, Yinbo Chen, Mohit Jain, and Xiaolong Wang.
\newblock Visual reinforcement learning with self-supervised 3d representations.
\newblock \emph{IEEE Robotics and Automation Letters}, 8\penalty0 (5):\penalty0 2890--2897, 2023.

\bibitem[Zeng et~al.(2021)Zeng, Florence, Tompson, Welker, Chien, Attarian, Armstrong, Krasin, Duong, Sindhwani, et~al.]{transporter}
Andy Zeng, Pete Florence, Jonathan Tompson, Stefan Welker, Jonathan Chien, Maria Attarian, Travis Armstrong, Ivan Krasin, Dan Duong, Vikas Sindhwani, et~al.
\newblock Transporter networks: Rearranging the visual world for robotic manipulation.
\newblock In \emph{Conference on Robot Learning}, pages 726--747. PMLR, 2021.

\bibitem[Zhao et~al.(2023)Zhao, Kumar, Levine, and Finn]{aloha}
Tony~Z Zhao, Vikash Kumar, Sergey Levine, and Chelsea Finn.
\newblock Learning fine-grained bimanual manipulation with low-cost hardware.
\newblock \emph{arXiv preprint arXiv:2304.13705}, 2023.

\bibitem[Zhu et~al.(2023)Zhu, Zhao, He, Zhong, Zhang, Yu, and Zhang]{zhu2023diffusion}
Zhengbang Zhu, Hanye Zhao, Haoran He, Yichao Zhong, Shenyu Zhang, Yong Yu, and Weinan Zhang.
\newblock Diffusion models for reinforcement learning: A survey.
\newblock \emph{arXiv preprint arXiv:2311.01223}, 2023.

\end{thebibliography}
}

\maketitlesupplementary
\begin{alphasection}

\section{Overview}

This supplementary document offers further information, results, and visualizations to complement the primary paper. Specifically, we encompass:

\begin{itemize}
\item Details on data collection;
\item Details on training and testing procedures;
\item Details on the design of evaluation levels;
\item Comparision to other data generation methods;
\item More ablation studies;
\item Additional details on the derivation of data augmentation for randomizing object poses.
\end{itemize}

\section{Implementation details}
In this section, we provide an overview of the data collection, training, and testing processes.

\subsection{Human Demonstration Collection}

The human play data is gathered through a teleoperation setup, where a human operator controls the system using a single real-sense camera in both the simulated and real environments. The entire trajectory is recorded at a rate of 30 frames per second, with each trajectory spanning approximately 20-30 seconds.

In the real-world setting, an additional real-sense camera is used to capture RGB images, which serve as the observations in the dataset. To ensure alignment between the simulated and real environments, we perform hand-eye calibration in the real world. This calibration process allows us to determine the relative position between the camera and the robot arm, enabling us to apply this transformation in the simulation.

\subsection{Real World Setup}
The system design for data collection is shown in Figure ~\ref{fig:system_setup}. As represented in the figure, the collection of human play data incorporates a human operator and a camera. The camera captures video footage at a frequency of 30 frames per second. Throughout the data collection process, the human operator interacts with the scene without any defined task objective. Instead, they interact freely with the environment, motivated by curiosity and the intent to observe intriguing behaviors.

In our experiments, human play data is collected by recording 30 seconds of uninterrupted interaction in each demonstration. This timeframe permits ample data to be gathered, yielding a rich and varied collection of behaviors for examination and study.

\begin{figure}[t]
  \centering
  \includegraphics[width=1\linewidth]{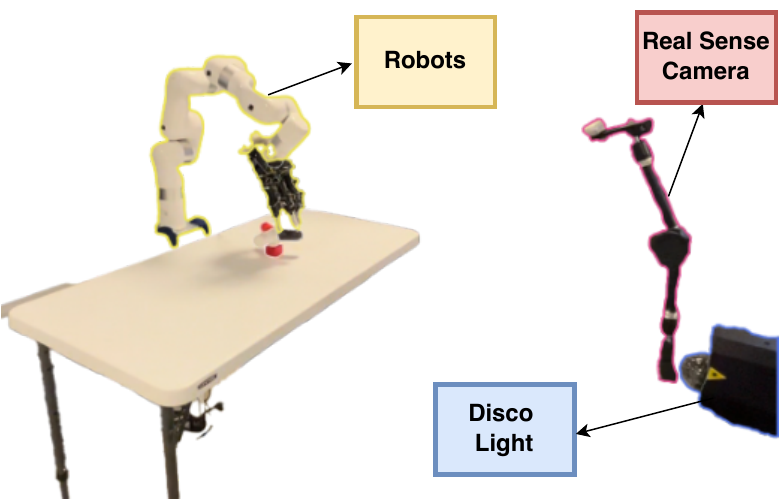}
   \caption{\textbf{Details of System Setups in Real World}}
   \label{fig:system_setup}
\end{figure}

\subsection{Policy Training}

We use a conditional VAE~\cite{cvae} for training on $100$ simulation demonstrations. The action chunking size was fixed at $50$, in line with the methodology adopted in~\cite{aloha}. Following the simulation data training, the model was fine-tuned with $15$ real-world demonstrations, using a smaller learning rate and distinct batch norms for the real-world data.

Hyperparameters related to policy learning are displayed in Table~\ref{tab:Policy}, whereas Table~\ref{tab:ACL} lists the hyperparameters pertinent to auto-curriculum learning.

To infuse diversity into the augmentation process, we have incorporated a randomness scale that ranges from $0$ to $10$ for each augmentation. In the context of auto-curriculum learning, this randomness scale progressively rises with a constant variance throughout each testing cycle.

In the course of auto-curriculum learning, the policy's performance is assessed across all four simulation levels, and the success rate is averaged. If the success rate falls below the success rate threshold, the increase in randomness scale is halted. This strategy aids in maintaining a balance between introducing randomness and ensuring the policy consistently accomplishes its tasks.

In summary, these hyperparameters and the evaluation procedure in auto-curriculum learning allow the policy to evolve and enhance over time, gradually escalating the randomness scale while preserving a satisfactory success rate.

\begin{table}[]
\centering
\begin{tabular}{ll}
\hline
Hyperparameter & Default \\ \hline
Batch Size & 128 \\
Num of Epochs & None \\
Finetuning Epochs & 3000 \\
Optimizer & AdamW \\
Learning Rate (LR) & 1e-5 \\
Finetuning LR & 1e-6 \\
Weight Decay & 1e-2 \\
Evaluation Frequency & 100 epochs \\
Encoder Layers & 4 \\
Decoder Layers & 7 \\
Heads & 8 \\
Feedforward Dimension & 3200 \\
Hidden Dimension & 256 \\
Chunk Size & 50 \\
Dropout & 0.1 \\ \hline
\end{tabular}
\caption{\textbf{Hyperparameters of Policy Network}}
\label{tab:Policy}
\end{table}

\begin{table}[]
\begin{tabular}{ll}
\hline
Hyper Parameters & Default \\ \hline
Test Cycles & 300 \\
Evaluation Freq & 100 epochs \\
Randomness Variance For Each Cycle & 0.2 \\
Success Rate Threshold & 15\% \\
Data Generation Rate Threshold & 30\% \\ \hline
\end{tabular}
\caption{\textbf{Hyperparameters for Auto Curriculum Learning}}
\label{tab:ACL}
\end{table}

\subsection{Policy Testing}
During the real-world testing phase, we perform both in-domain and out-of-domain tests to evaluate the performance of the model. For out-of-domain tests, we significantly randomize the positions of objects, consciously choosing locations not included in the original data. This step guarantees that the model is examined in unfamiliar situations, evaluating its capacity to generalize and adjust to novel object arrangements.

Moreover, to introduce visual disruptions and test the robustness of the model, we incorporate a disco light. The disco light generates visual disturbances and adds an extra layer of complexity to the test environment. This approach enables us to assess the model's resilience in dealing with unexpected visual inputs and its ability to sustain performance amidst such disruptions.

In the concluding stage, we evaluate the policy's ability to generalize across a range of objects, as illustrated in Figure~\ref{fig:object_set}. To carry out this generalizability test, we enhance the initial $100$ simulation demonstrations by introducing $10$ unique objects (adding $10$ additional demonstrations for each object), and then re-run our pipeline. For the pick and place task in the real-world setting, we collected 15 demonstrations involving three different objects (with five demonstrations performed for each object within the red frame). For the rotating task, the real-world dataset includes only one object, identical to the original testing case.

By conducting these assessments and incorporating a variety of objects, we aim to evaluate the policy's adaptability and performance in diverse situations, ensuring its robustness and flexibility. A selection of demos is displayed in Figure~\ref{fig:test_diverse_objects}.

\subsection{Details of Simulation Evaluation Level designs}
In the {Pick \& Place} and {Pour} tasks, we have defined different levels to introduce varying degrees of randomness:
\begin{itemize}
\item Level 1 signifies the original domain and encompasses slight randomization of the pose of the manipulated objects, including the end-effector pose and orientation.
\item Level 2 includes randomization of lighting and texture.
\item Level 3 incorporates minimal randomization of the target objects (plate in pick place, bowl in pouring).
\item Level 4 escalates the randomness scale of both the manipulated and target objects.
\end{itemize}
For {Rotate} task since there is only one object, things are different for the {Rotate} task: 
\begin{itemize}
\item Level 1 is set as randomizing the orientation of the objects, which is also the original domain. 
\item Level 2 is the same as pick place and pouring tasks.
\item Level 3 is adding the randomization of the end-effector pose of the manipulated objects. 
\item Level 4 increases the randomness scale of the manipulated objects. 
\end{itemize} 
Below are the defined randomness parameters for each task, with all numbers listed in international units if without a statement. In our settings, the position (0,0) represents the center of the table.

\noindent \textbf{Level Design for Pick and Place}
\begin{itemize}
\item Random Manipulated Object Pose with a small scale:
    \begin{itemize}
        \item The x-coordinate of the Manipulated Object ranges from -0.1 to 0.1.
        \item The y-coordinate of the Manipulated Object ranges from 0.2 to 0.3.
        \item The Manipulated Object's z-axis Euler degree ranges from 80 to 90.
    \end{itemize}
\item Random Light and Texture(The randomness scale here is fixed to be 2):
    \begin{itemize}
        \item The direction of the light is constrained within a circular range. The radius of this circle spans from 0.5 to the randomness scale * 0.1. 
        \item To determine the color of each channel for the lights, a uniform sampling approach is employed. This involves selecting a value within the range [default color of that channel - randomness scale * 0.1, default color of that channel + randomness scale * 0.1].
        \item The ground color and sky color of the environment map are randomized in the same way as lights.
    \end{itemize}
\item Random Target Object Pose with a small scale: 
    \begin{itemize}
         \item The x-coordinate of the Target Object ranges from -0.1 to 0.1.
        \item The y-coordinate of the Target Object ranges from -0.3 to -0.1.
    \end{itemize}
\item Random Manipulated and Target Object Pose with a Large scale: 
    \begin{itemize}
        \item The x-coordinate of the Manipulated Object ranges from -0.2 to 0.2.
        \item The y-coordinate of the Manipulated Object ranges from 0.1 to 0.3.
        \item The Manipulated Object's z-axis Euler degree ranges from 70 to 90.
        \item The Manipulated Object's z-axis Euler degree ranges from 80 to 90.
        \item The x-coordinate of the Target Object ranges from -0.2 to 0.2.
        \item The y-coordinate of the Target Object ranges from -0.3 to 0.
    \end{itemize}
\end{itemize}

\noindent \textbf{Level Design for Pour}
\begin{itemize}
\item Random Manipulated Object Pose with a small scale:
    \begin{itemize}
        \item The x-coordinate of the Manipulated Object ranges from -0.1 to 0.1.
        \item The y-coordinate of the Manipulated Object ranges from -0.2 to -0.1.
        \item The Manipulated Object's z-axis Euler degree ranges from 0 to 179.
    \end{itemize}
\item Random Light and Texture(The randomness scale here is fixed to be 2): same as pick and place.
\item Random Target Object Pose with a small scale: 
    \begin{itemize}
         \item The x-coordinate of the Target Object ranges from -0.1 to 0.1.
        \item The y-coordinate of the Target Object ranges from 0.2 to 0.3.
    \end{itemize}
\item Random Manipulated and Target Object Pose with a Large scale: 
    \begin{itemize}
        \item The x-coordinate of the Manipulated Object ranges from -0.1 to 0.15.
        \item The y-coordinate of the Manipulated Object ranges from -0.3 to 0.
        \item The x-coordinate of the Target Object ranges from -0.2 to 0.2.
        \item The y-coordinate of the Target Object ranges from 0.2 to 0.4.
        \item The Manipulated Object's z-axis Euler degree ranges from 0 to 359.
    \end{itemize}
\end{itemize}

\noindent \textbf{Level Design Rotate}
\begin{itemize}
\item Random Manipulated Object Pose with a small scale:
    \begin{itemize}
        \item The Manipulated Object's z-axis Euler degree ranges from 0 to 30.
    \end{itemize}
\item Random Light and Texture(The randomness scale here is fixed to be 2): same as pick and place.
\item Random Manipulated Object Pose with a small scale: 
    \begin{itemize}
         \item The x-coordinate of the Target Object ranges from -0.1 to 0.1.
        \item The y-coordinate of the Target Object ranges from -0.15 to 0.15.
    \end{itemize}
\item Random Manipulated Pose with a Large scale: 
    \begin{itemize}
        \item The x-coordinate of the Manipulated Object ranges from -0.2 to 0.2.
        \item The y-coordinate of the Manipulated Object ranges from -0.3 to 0.3.
        \item The Manipulated Object's z-axis Euler degree ranges from 0 to 60.
    \end{itemize}
\end{itemize}

\begin{figure}[t]
  \centering
  \includegraphics[width=1\linewidth]{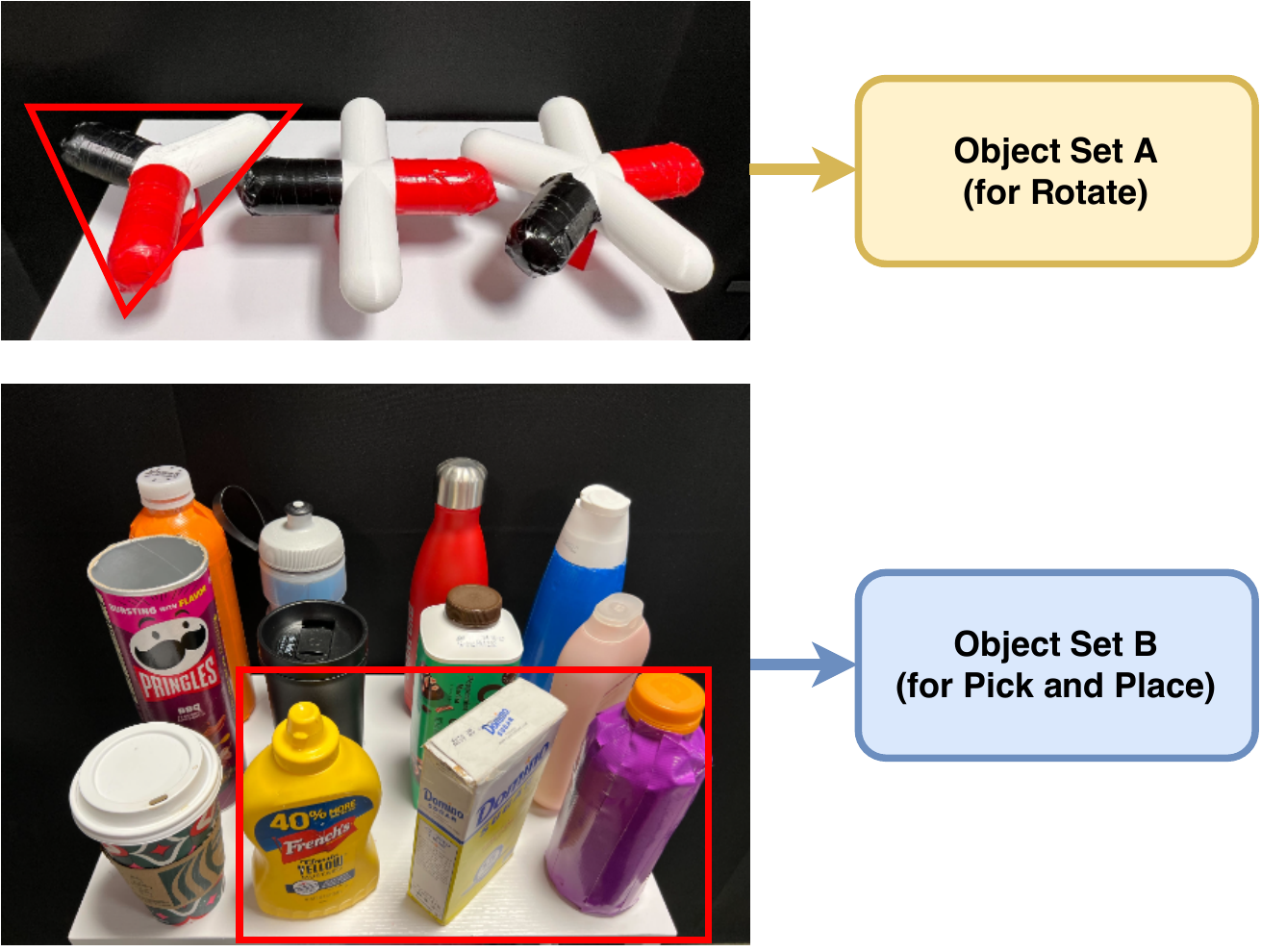}
   \caption{\textbf{Object Sets in Real World.} The objects located within the red frame are allocated for training, while the remaining objects are set aside for testing on previously unseen objects.}
   \label{fig:object_set}
\end{figure}

\begin{figure}[t]
  \centering
  \includegraphics[width=1\linewidth]{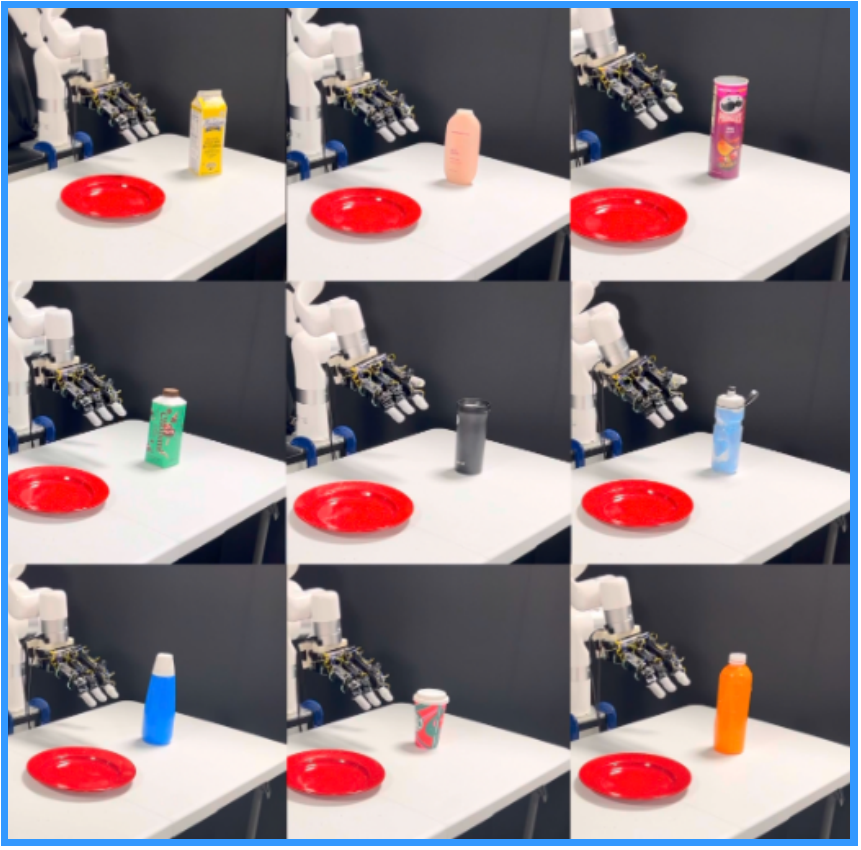}
   \caption{\textbf{Pick and Place Evaluation on diverse real-world objects.}}
   \label{fig:test_diverse_objects}
\end{figure}

\section{More Experimental Results}
\subsection{Comparision of Data Generation Method}
To test our data augmentation approach's effectiveness, we pretrained using simulation data augmented by MimicGen~\cite{mimicgen} and then fine-tuned with real-world teleoperation data, a sim2real transfer not included in the original MimicGen framework.
As shown in Figure~\ref{tab:mimicgen}, MimicGen adds an interpolated trajectory (in purple) to new object poses, potentially causing abrupt transitions. Our method, however, seamlessly integrates the entire sequence, resulting in more fluid motion. The imitation learning policy trained with our data thus outperforms others, as evidenced by the improved results in simulation and reality shown in the table.

\begin{table}[t]
\resizebox{\columnwidth}{!}{%
\begin{tabular}{l|cccc|ccc}
  & \multicolumn{4}{c|}{\textbf{Simulation}} & \multicolumn{3}{c}{\textbf{Real World}} \\
 & Level 1 & \multicolumn{1}{l}{Level 2} & \multicolumn{1}{l}{Level 3} & \multicolumn{1}{l|}{Level 4} & \begin{tabular}[c]{@{}c@{}}In \\ Domain\end{tabular} & \begin{tabular}[c]{@{}c@{}}Out of \\ Position\end{tabular} & \begin{tabular}[c]{@{}c@{}}Random \\ Light\end{tabular}  \\ \hline
MimicGen & 75.5\% & 49\% & 19.5\% & 14\% & 2/20 & 1/20 & 5/20 \\
Ours & \textbf{80\%} & \textbf{61\%} & \textbf{43.5\%} & \textbf{57\%} & \textbf{7/20} & \textbf{6/20} & \textbf{8/20}
\end{tabular}%
}
\caption{\textbf{Comparision to MimicGen}. We compare our data augmentation method with the one used in MimicGen on the Pick and Place task. For real-world experiments, we fine-tuned it with the same real-world data as other methods. }
\label{tab:mimicgen}
\end{table}

\begin{figure}[h]
  \vspace{-3mm}
  \centering
  \includegraphics[width=1\linewidth]{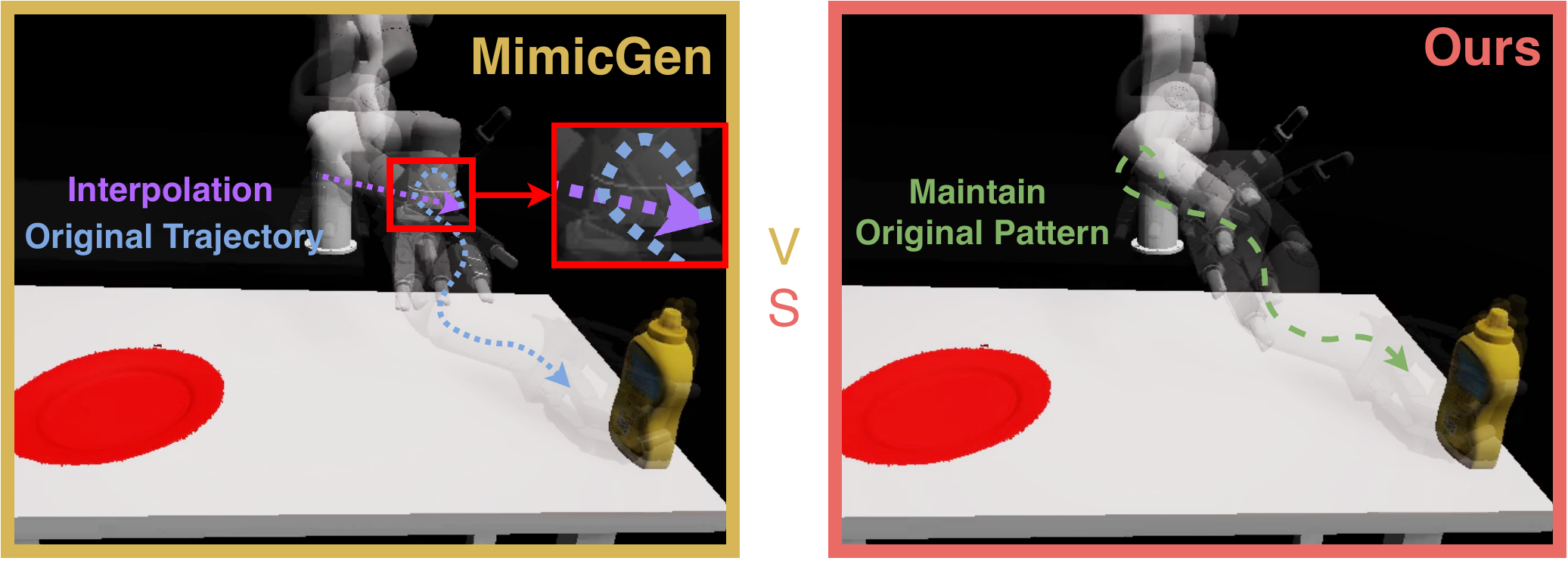}
   \label{fig:mimicgen}
\vspace{-6mm}
\end{figure}

\subsection{Ablation on Action Aggregation}
\begin{figure}[t]
  \centering  \includegraphics[width=1\linewidth]{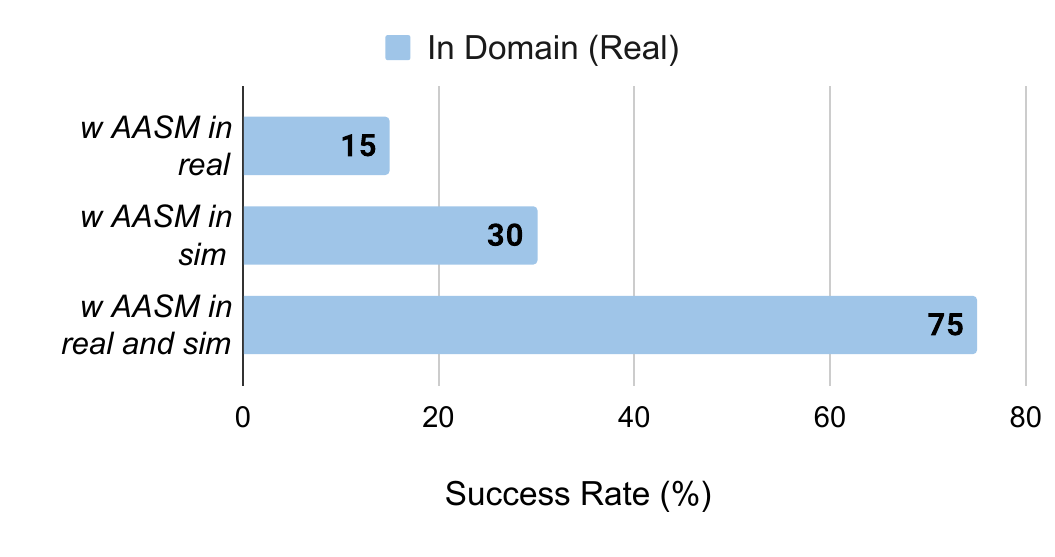}
   \caption{\textbf{Success Rate  on Action Aggregation} }
   \label{fig:aasm_real}
\end{figure}

\begin{figure}[t]
  \centering
  \includegraphics[width=1\linewidth]{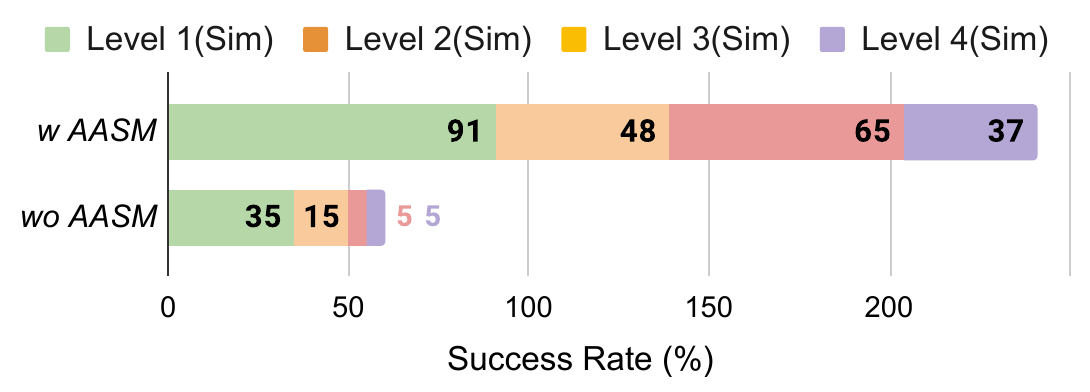}
   \caption{\textbf{Ablation on Action Aggregation with Small Motion in Simulation.} Success rate evaluate in simulator when the policy is trained on dataset with and without action aggregation.}
   \label{fig:aasm_sim}
\end{figure}

As illustrated in Figure~\ref{fig:aasm_real}, the use of action aggregation with Small Motion extends beyond its advantages in imitation learning within a single domain. It also functions as an effective instrument in closing the gap between simulated and real environments. As illustrated in Figure~\ref{fig:aasm_sim}, the success rate of the policy becomes more pronounced as the level of difficulty escalates. This suggests that action aggregation becomes increasingly beneficial for more challenging tasks.

\begin{figure}[t]
  \centering
  \includegraphics[width=1\linewidth]{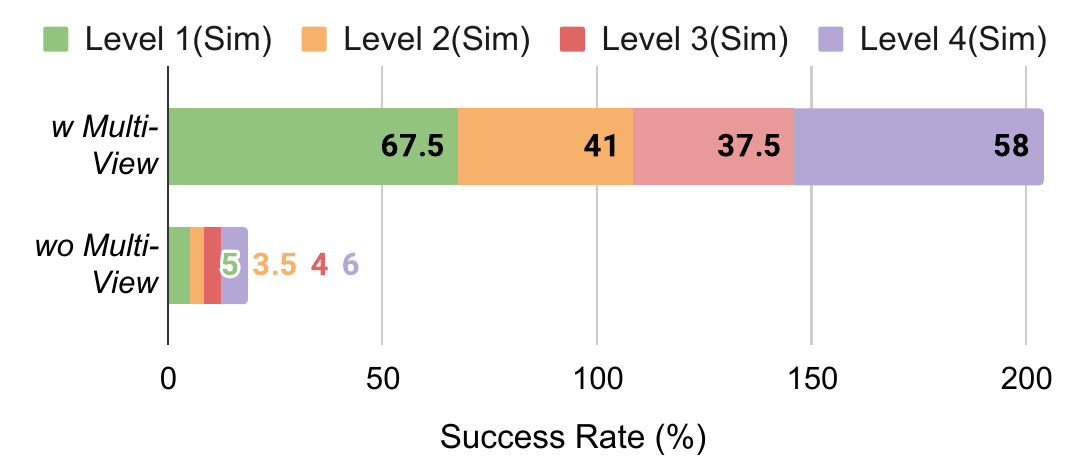}
   \caption{\textbf{Ablation on Multi-View Augmentaion}}
   \label{fig:mul_view}
\end{figure}

\subsection{Ablation on Novel Camera Views}
As demonstrated in Figure~\ref{fig:mul_view}, the application of random camera views augmentation improves the policy's robustness, particularly in situations involving camera view changes. In this method, all levels remain consistent while minor alterations to the camera view are incorporated.

These changes involve a combined rotation along the y-axis and z-axis, plus a slight shift in the x, y, and z directions. The rotation Euler angle is sampled within the range of $[-15, 15]$, enabling managed variation in the camera's alignment. Additionally, the translation is sampled within the range of $[-0.05m, 0.05m]$, allowing minor adjustments in the camera's placement.

By integrating these alterations, the multi-view augmentation method introduces realistic variations in camera perspectives, thereby enhancing the policy's resilience to changes in the viewpoint. This strategy boosts the model's capability to adapt and perform efficiently, even when confronted with varied camera angles and positions.

\subsection{Ablation on Kinematics Augmentation}
We ablate the augmentation methods with or without sensitivity analysis (in contrast, simply relocating the end-effector to a new pose). We test both methods in an environment where object poses are extensively randomized, to verify the effectiveness of these two kinematic augmentation approaches.
Figure~\ref{fig:Sensitive_Analysis} illustrates the success rate during training using datasets generated via these distinct augmentation strategies. Our findings demonstrate that, through the application of our proposed techniques, the model demonstrates consistent improvement over all three manipulation tasks. These outcomes underscore the efficacy of incorporating sensitivity analysis into pose data augmentation. 

\begin{figure}[h]
  \vspace{-3mm}
  \centering
  \includegraphics[width=1\linewidth]{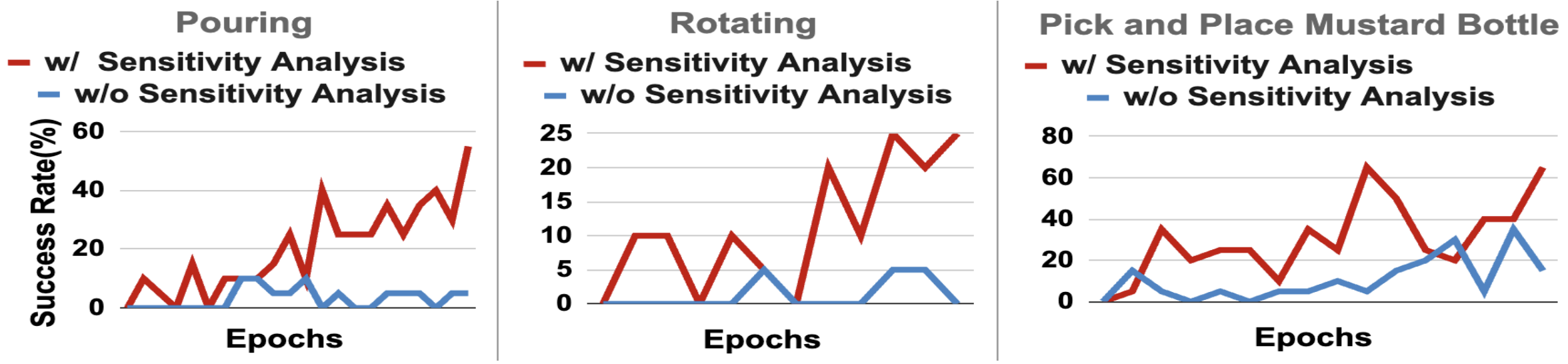}
   \caption{\textbf{Augmentation Comparison} We apply kinematic augmentation to one selected original demo, adapting it to a new position utilizing both the MimicGen method and ours.}
   \label{fig:Sensitive_Analysis}
   \vspace{-7mm}
\end{figure}

\section{Derivation of Data Augmentation}

In this section, we delve deeper into the derivation of the formula applied in the data augmentation of random object pose. We reproduce Equation 2 from the main paper and provide a detailed explanation:

\begin{equation}
    \begin{split}
    \overline{\psi}_{seg_j} &= \frac{\psi_{seg_j}}{\sum_{j=1}^{M} \psi_{seg_j}}, \quad \forall {seg_j} \\
    \Delta T_j &= exp(\overline{\psi}_{seg_j} \log (\Delta T)/K) \\
    a_{i}^{new} &= a_{i} f_i(\Delta T_{j})   
    \end{split}
\end{equation}

The first line of the equation normalizes the robustness score computed from Equation 1 in the main paper, ensuring that the sum of all scores equals $1$. This parameter can be interpreted as a weight for each action chunk, symbolizing the proportion of modification each chunk should undertake to guide the robot to the new pose.

The second line calculates the relative pose modification for each step in chunk $j$. There are $K$ steps in chunk $j$, and each step is allocated the same quantity of modification within the same chunk. Here, $log()$ maps the $SE(3)$ Lie Group to its $se(3)$ Lie algebra, where $exp$ is its inverse, mapping $se(3)$ back to $SE(3)$.

The third line of this equation computes the new action based on the pose modification. Here, $f_i$ is a similarity transformation in the $SE(3)$ space that transitions the motion from the world frame to the current end-effector frame. We now provide a detailed derivation of $f_i$. Since $\Delta T = T_{W}^{O_{new}} (T_{W}^{O_{old}})^{-1} = T_{O_{old}}^{O_{new}}$, this relative pose change is a representation in the old object pose frame. To use this pose transformation to modify the action, we need to transform this relative pose into the frame corresponding to the action, which is the frame of the current end-effector pose. Considering the similarity transformation $T_B^A X (T_B^A)^{-1}$, which transforms a $SE(3)$ motion $X$ represented in frame $B$ to frame $A$, $f_i$ can be derived as $f_i(\Delta T) = T_{R_i}^{O_{old}} \Delta T (T_{R_i}^{O_{old}})^{-1}$, where $T_{R_i}$ is the robot end-effector pose in frame $i$.

\end{alphasection}

\end{document}